\documentclass[10pt,journal,compsoc]{IEEEtran}

\ifCLASSOPTIONcompsoc
  \usepackage[nocompress]{cite}
\else
  \usepackage{cite}
\fi
\ifCLASSINFOpdf
\else
\fi

\usepackage{subfigure}
\usepackage{graphicx}
\usepackage{booktabs} 
\usepackage{adjustbox}
\usepackage{subcaption}
\usepackage{multirow}

\usepackage{algorithm}

\usepackage{amsmath}
\usepackage{amssymb}
\usepackage{xcolor}
\usepackage{breqn,xspace}
\usepackage{bm}
\usepackage{lipsum,multicol}
\usepackage{makecell}
\usepackage{booktabs}
\usepackage{algorithm}
\usepackage{amsmath}

\usepackage{algpseudocode}
\usepackage{amsmath}
\usepackage{diagbox}
\allowdisplaybreaks[2]

\ifodd 0
\definecolor{darkgreen}{RGB}{0,200,0}
\newcommand{\rev}[1]{{\color{blue}#1}} 
\else
\newcommand{\rev}[1]{#1}
\fi

\begin{document}
%
\newcommand{\name}{FedSN\xspace}

\title{{\name: A Federated Learning Framework over Heterogeneous LEO Satellite Networks}}
%
%
%
%


\author{Zheng Lin, Zhe Chen,~\IEEEmembership{Member,~IEEE}, Zihan Fang,   Xianhao Chen,~\IEEEmembership{Member,~IEEE}, \\Xiong Wang,~\IEEEmembership{Member,~IEEE} and Yue Gao,~\IEEEmembership{Fellow,~IEEE}

\IEEEcompsocitemizethanks{
\IEEEcompsocthanksitem
This work was supported by the National Key R\&D Program of China under Grant No. 2023YFE0116600. The work of X. Chen was supported in part by the Research Grants Council of Hong Kong under Grant 27213824 and in part by HKU IDS Research Seed Fund under Grant IDS-RSF2023-0012.
\IEEEcompsocthanksitem
Z. Lin, Z. Chen, Z. Fang, X. Wang and Y. Gao are with the Institute of Space Internet, Fudan University, Shanghai 200438, China, and the School of Computer Science, Fudan University, Shanghai 200438, China (e-mail: zlin20@fudan.edu.cn; zhechen@fudan.edu.cn; zhfang19@fudan.edu.cn; wangxiong@fudan.edu.cn; gao.yue@fudan.edu.cn ).
\IEEEcompsocthanksitem
X. Chen is with the Department of Electrical and Electronic Engineering, University of Hong Kong, Pok Fu Lam, Hong Kong, China (e-mail: xchen@eee.hku.hk).
}

\thanks{\textit{\quad\quad\!\!(Corresponding author: Yue Gao)}}}

%
%

\markboth{Journal of \LaTeX\ Class Files,~Vol.~14, No.~8, August~2015}%
{Shell \MakeLowercase{\textit{et al.}}: Bare Advanced Demo of IEEEtran.cls for IEEE Computer Society Journals}
%



\IEEEtitleabstractindextext{%
\begin{abstract}
Recently, a large number of Low Earth Orbit (LEO) satellites have been launched and deployed successfully in space. Due to multimodal sensors equipped by the LEO satellites, they serve not only for communication but also for various machine learning applications. However, the ground station (GS) may be incapable of downloading such a large volume of raw sensing data for centralized model training due to the limited contact time with LEO satellites~(e.g.~5 minutes). Therefore, \textit{federated learning} (FL) has emerged as the promising solution to address this problem via on-device training. Unfortunately, enabling FL on LEO satellites still face three critical challenges: i) heterogeneous computing and memory capabilities, ii) limited downlink/uplink rate,  and iii) model staleness. To this end, we propose \textbf{\name} as a general FL framework to tackle the above challenges. Specifically, we first present a novel sub-structure scheme to enable heterogeneous local model training considering different computing, memory, and communication constraints on LEO satellites. Additionally, we propose a pseudo-synchronous model aggregation strategy to dynamically schedule model aggregation for compensating model staleness. Extensive experiments with real-world satellite data demonstrate that \name framework achieves higher accuracy, lower computing, and communication overhead than the state-of-the-art benchmarks.
\end{abstract}

}

\maketitle

\IEEEdisplaynontitleabstractindextext

%
\IEEEpeerreviewmaketitle


\section{Introduction}

The significant progress made in satellite technology has resulted in a sharp reduction in the expenses associated with launching satellites. This has encouraged an enormous expansion in the number of satellites orbiting the Earth at low altitudes. Several corporations have pledged to launch thousands of Low Earth Orbit~(LEO) satellites over the next few years~\cite{mark2018tech}. For example, SpaceX~\cite{ahmmed2022digital} plans to launch about 42,000 LEO satellites, and different from geostationary~(GEO) satellites, LEO satellite networks always construct meta-constellations for communications. More importantly, LEO satellites collect large amounts of sensor data~\cite{chen2022robust,wu2024accelerating,yuan2023graph}, such as spectrum and earth imageries, which can be leveraged to resolve many global challenges such as food safety~\cite{aragon2018cubesats}, disease spread~\cite{franch2020spatial}, and climate change~\cite{shukla2021enhancing} with the power of deep learning.

It is relatively easy to train deep learning models using plentiful data in terrestrial networks in a timely and synchronous fashion~\cite{letaief2021edge,chen2021rf,lyu2023optimal,fang2024automated,lin2024splitlora}.
However, such conditions do not hold in the space networks, since conventional centralized learning frameworks over satellite networks are significantly hindered by the limited contact time between LEO satellites and a ground station~(GS). We face a fundamental problem that datasets are often too large to download to a GS~\cite{so2022fedspace}. Moreover, the satellite nodes may not be willing to share the data with GSs due to commercial interests and data ownership concerns if they belong to different parties~\cite{lin2023efficient}. Fortunately, federated learning~(FL)~\cite{mcmahan2017communication} is a new paradigm for distributed training technology on massive participating edge devices. FL only needs a central server to coordinate all participating edge devices, and aggregate their local model parameters~\cite{chen2020joint,xu2020client,zhang2024fedac,lin2024adaptsfl}. In this way, it does not only protect data privacy, but also reduce communication overhead between a central server and edge devices.

\begin{figure}[t!]
\centering
\includegraphics[width=7.2 cm]{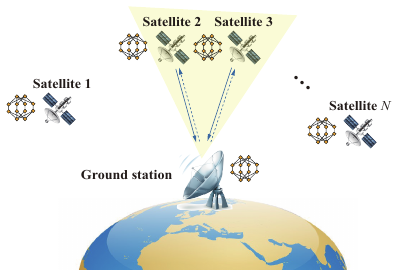}
\vspace{-0.2cm}
\caption{A scenario of FL over LEO satellite networks.
}
\label{fig:Fed_spa_scenario}
\end{figure}

Generally, for FL over LEO satellite networks,  satellites train local models using their own datasets, and GS as a central server aggregates local models of satellites as shown in Fig.~\ref{fig:Fed_spa_scenario}\footnote{In this case, we do not consider inter-satellite link, since a totally distributed FL in space faces more limitations such as energy, communication, the complexity of architecture, etc. Moreover, even for Starlink, there is no evidence that inter-satellite link works in current LEO satellite networks~\cite{handley2019using,so2022fedspace}.}.
However, current FL frameworks, such as FedAvg~\cite{mcmahan2017communication}, cannot work effectively for satellite networks, since LEO satellites always have limited connectivity with GS~(e.g., 5 minutes contact time), leading to synchronous FL update failure. 
Recently, several researchers have drawn more attention to designing asynchronous FL frameworks for LEO satellite networks~\cite{razmi2022ground,so2022fedspace,wu2023fedgsm,elmahallawy2022asyncfleo,razmi2022scheduling}. Razmi~\textit{et al.}~\cite{razmi2022ground} propose a predictive satellites-GS asynchronous FL framework to aggregate local updates, but that asynchronous FL for LEO satellites cannot tackle staleness well. In view of this, there are some research works~\cite{wu2023fedgsm,elmahallawy2022asyncfleo,razmi2022scheduling} study different staleness weighting functions to solve it in LEO satellite networks. All of those works only consider time-dependent weighting functions for aggregating previous and current models without accounting for discrepancies between the two models. We will illustrate how the discrepancies impact on model aggregation in Sec.~\ref{sec:background_motivation}.

Apparently, to design a general FL framework for LEO satellite networks is a non-trivial task, and we need to solve three practical challenges. First of all, LEO satellites have heterogeneous computing and memory capabilities. For example, various versions of satellites carry varied computing resources. Even satellites with the identical computing capabilities may execute other computing tasks, leading to significantly varied idle resources. Second, while downlink~(from satellite to GS) data rate is high, the low uplink data rate becomes a bottleneck for FL model aggregation. We will do measurement studies in Sec.~\ref{sec:background_motivation} to verify this statement. Last but not least,  model staleness compensation needs to be reconsidered, since the current FL works for LEO satellites only consider single orbit and time-dependent compensation, but not discrepancies between different local models.


To tackle the above challenges, we propose a general FL framework for LEO satellite networks, named \textit{\name}. Our \name incoporates two components that are \textit{sub-structure scheme} and \textit{pseudo-synchronous model  aggregation}. For sub-structure scheme, we propose a unified constrained budget to map the computing and memory resources, and uplink rate into a utility value. {According to the minimum constrained budget chosen from distinct satellites' demands\footnote{{Determining the basic sub-models with the minimum budget of participating satellites ensures that each sub-structure model can be trained across all satellites.}}, we customize basic sub-structure models, and split a global model into several basic sub-structure models.} Thereafter, we select sub-structure combinations, and assign them to satellites. While the satellite obtains model combination, the satellite trains it locally. Following standard FL procedure, all basic sub-structure models are aggregated in a GS after training.
However, sub-structure scheme brings a new challenge for model staleness, and hence, for pseudo-synchronous model  aggregation, we design a similarity evaluation function to portray the extent of model staleness, and propose buffer-based model aggregation scheme to compensate the model staleness. Hereby, pseudo-synchronous model aggregation strategy expedites model convergence and improves the test accuracy. All in all, resorting to our two approaches, \name overcomes the aforementioned challenges of deploying FL over LEO satellite networks.
We summarize our key contributions in the following.
\begin{itemize}
    \item {We propose a unified constrained budget to map the computing and memory resources, and uplink rate into a utility value. Based on this, we design an FL sub-structure scheme to handle heterogeneous computing and memory resources and limited downlink/uplink rate for LEO satellite networks to achieve superior training performance.}
    \item Different from state-of-the-art FL approaches for LEO satellite networks, we design a novel aggregation scheme to improve the training accuracy of FL by taking model discrepancy into account. 
    \item We empirically demonstrate the effectiveness of our \name using real-world LEO satellite networks and two types of datasets~(space modulation recognition and remote sensing image classification). The results illustrate that \name outperforms the state-of-the-art FL frameworks.
\end{itemize}

This paper is organized as follows. Sec.~\ref{sec:background_motivation} motivates the design of \name\ by revealing the limitations in current LEO satellite networks. Sec.~\ref{sec:system_design} presents the system design of \name. Sec.~\ref{sec:implementation} introduces system implementation, and experiment setup, followed by performance evaluation in Sec.~\ref{sec:evaluation}. Related works and technical limitations are discussed in Sec.~\ref{sec:related_work}. Finally, conclusions and future remarks are presented in Sec.~\ref{sec: conclusion}.

\section{Motivation and Background} \label{sec:background_motivation}
In this section, for better motivating the design of \name, motivating measurement studies are provided by us to concretely demonstrate the challenges faced by deploying FL over LEO satellite networks.

\subsection{Heterogeneous Computing and Memory Budgets} \label{ssec:hrtero_compute_memory}

\begin{figure}[t]
\vspace{-.5ex}
\setlength\abovecaptionskip{3pt}
\centering
\subfigure[Computing.]{
    \includegraphics[height=2.45cm,width=4cm]{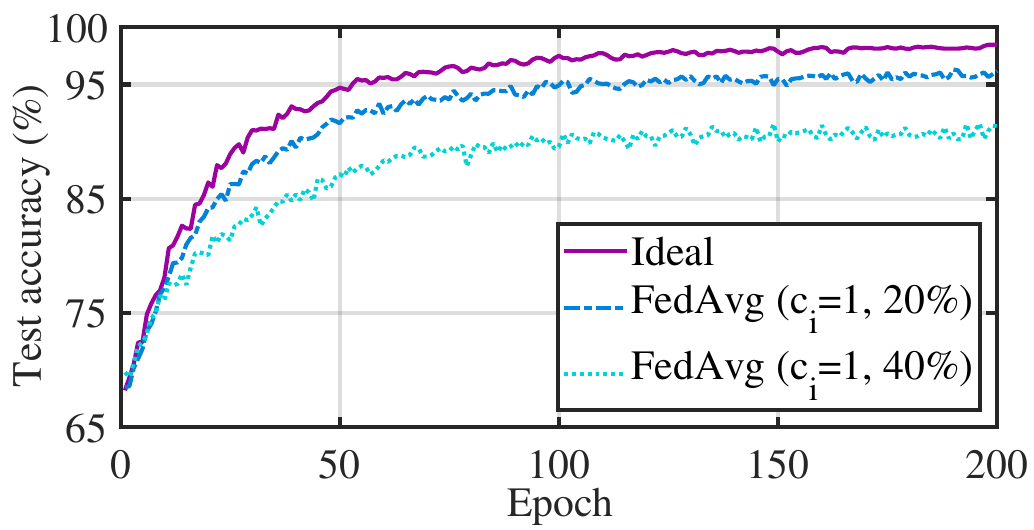}
    \label{sfig:hetero_compute}
}
\subfigure[Memory.]{
    \includegraphics[height=2.4cm,width=4cm]{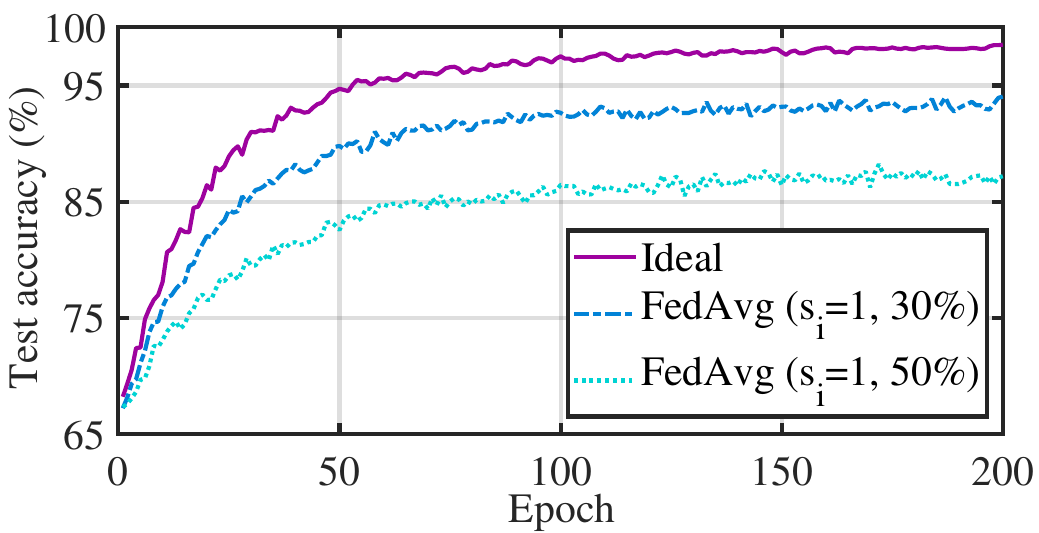}
    \label{sfig:hetero_memory}
}
    \caption{ The impact of heterogeneous computing and memory resources on FL. Fig.~\ref{sfig:hetero_compute} and Fig.~\ref{sfig:hetero_memory} show the performance for test accuracy versus under-training satellite rates under computing and memory constraints. The experiment is conducted on the GBSense~\cite{gbsense} under the IID setting using VGG-16~\cite{simonyan2014very}. }
    \label{fig:hetero_resource}
    \vspace{-2ex}
\end{figure}

For the deployment of FL for satellite network, the existing FL frameworks~\cite{so2022fedspace,wu2023fedgsm,elmahallawy2022asyncfleo,razmi2022scheduling,razmi2022ground} assume that each satellite can offer enough on-device resources~(computing power and memory size) to train a local model, and hence, a GS always can aggregate all trained local models to obtain good performance. However, in practice, various satellites have different on-device resources, and their allocated resources for FL training may be drastically changed during run-time, since those allocated resources depend on how the satellite's running programs prioritize on-device resource allocation. Hereby, due to heterogeneous on-device resources of different satellites, FL aggregation in GS may suffer from under-training local models under some satellites.

To better understand how heterogeneous on-device resources affect FL performance, we set up two experiments for run-time computing power and memory size. We utilize FedAvg, which is one of the most popular FL instantiations to study its performance under two experiments, and a well-known VGG-16~\cite{simonyan2014very} widely applied in many applications~(e.g., object detection~\cite{haque2019object}) as local model. We train local models using the space modulation recognition dataset GBSense~\cite{gbsense}. We set the total number of satellites as 10, and emulate each satellite using a virtual machine. We also randomly assign computing power and memory sizes to virtual machines and set their resource budget to emulate heterogeneous satellites.

\begin{figure}[t]
\setlength\abovecaptionskip{3pt}
\centering
\subfigure[A GS of Starlink.]{
    \includegraphics[height=2.4cm,width=4cm]{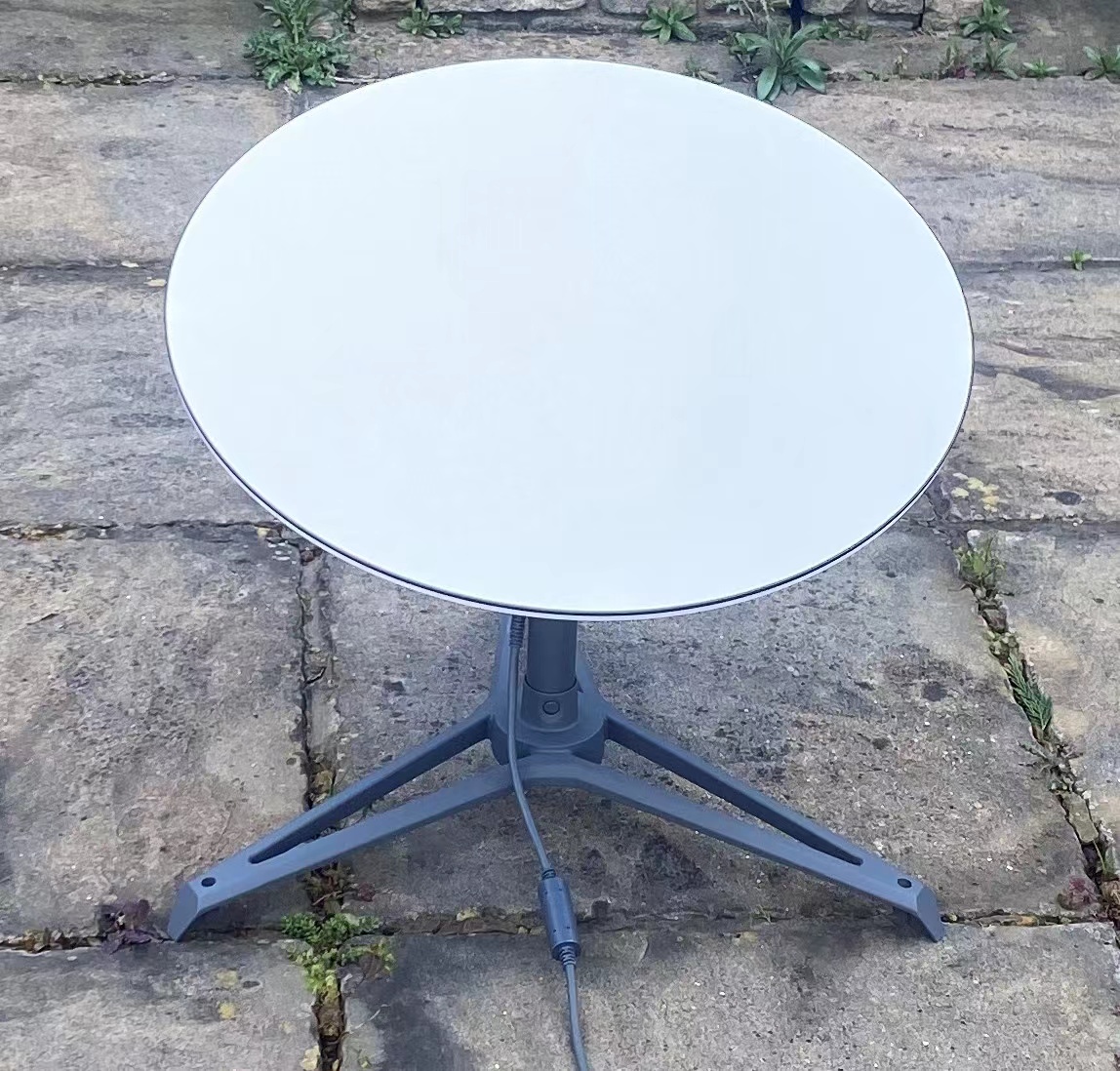}
    \label{sfig:exStarlink}
}
\subfigure[Experimental setup.]{
    \includegraphics[height=2.4cm,width=4cm]{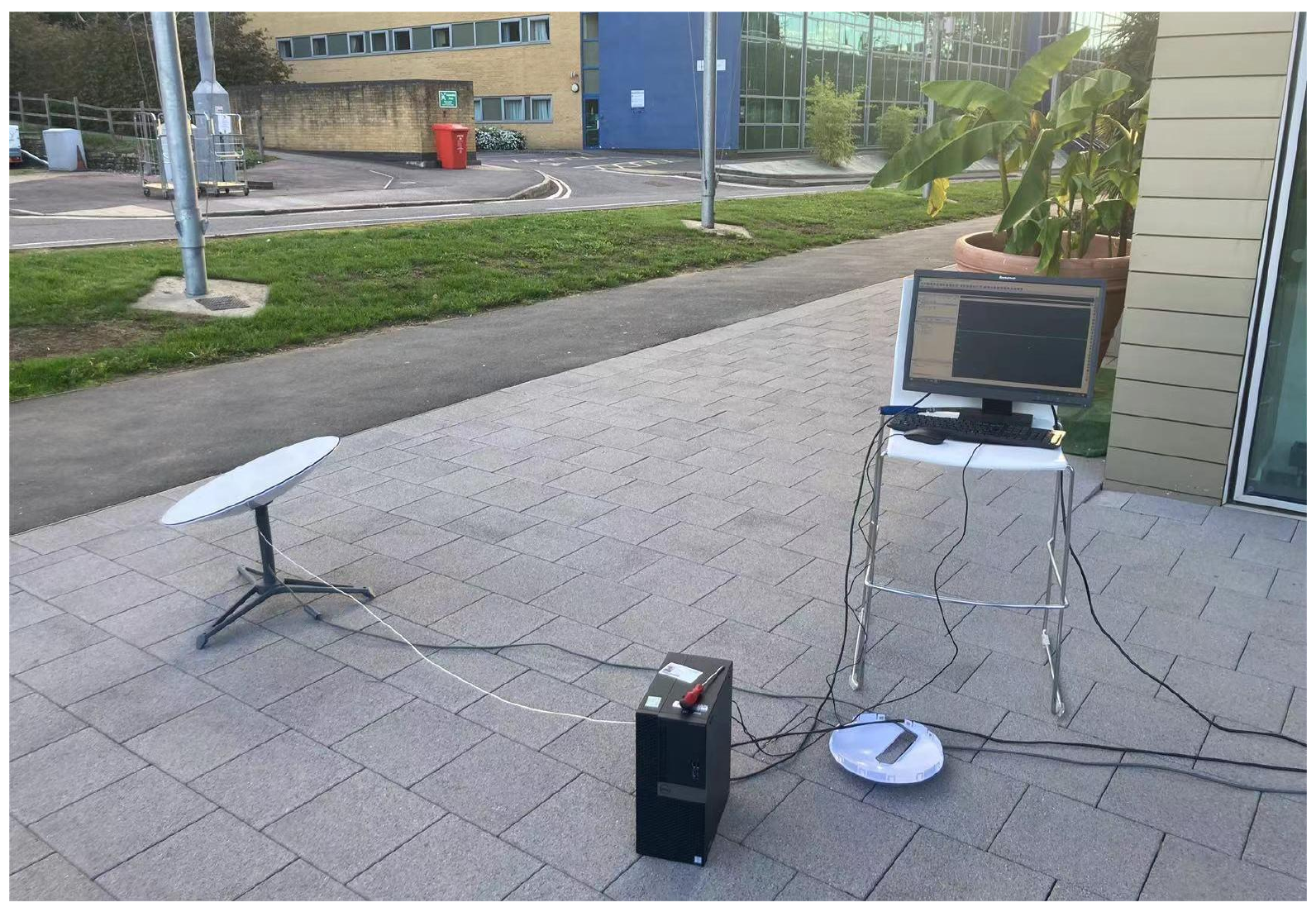}
    \label{sfig:gsStarlink}
}
\subfigure[CDF of uplink and downlink rate.]{
    \includegraphics[height=2.4cm,width=4cm]{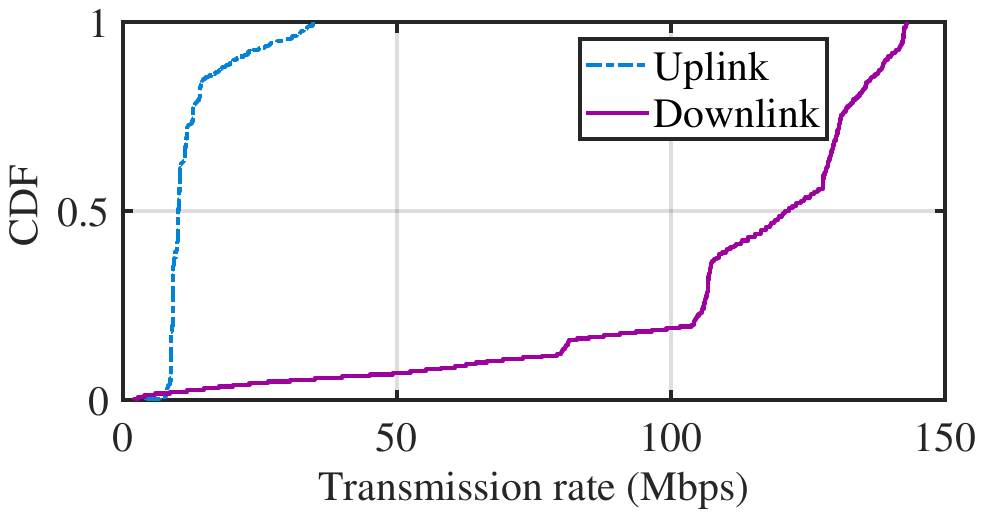}
    \label{sfig:data_rate}
}
\subfigure[Local models updating failure.]{
    \includegraphics[height=2.4cm,width=4cm]{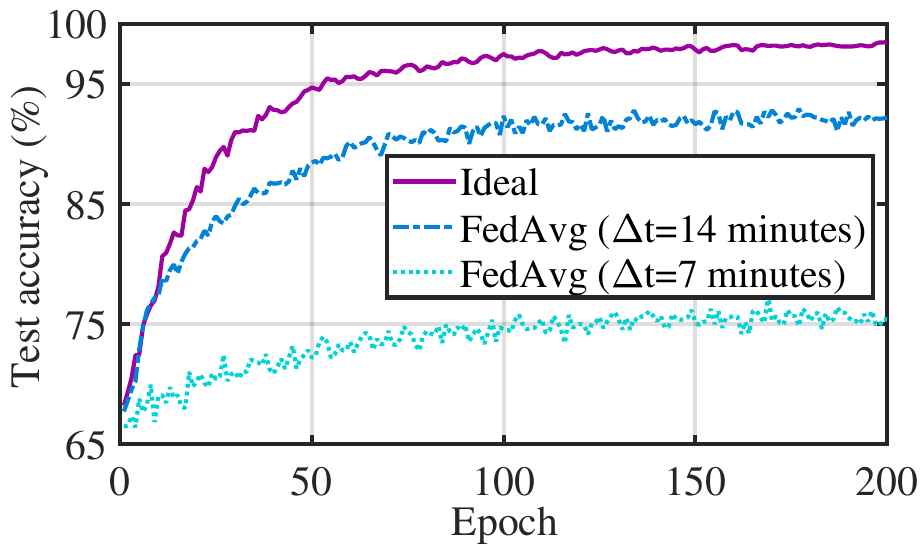}
    \label{sfig:update_fail}
}
    \caption{Uplink communication between LEO satellite and GS becomes a major bottleneck for FL. Fig.~\ref{sfig:exStarlink} and {Fig.~\ref{sfig:gsStarlink} show Starlink's GS and experimental setup for average uplink and downlink rate measurements.} Fig.~\ref{sfig:data_rate} presents the CDF of the uplink and downlink rates. {Fig.~\ref{sfig:update_fail} illustrates the performance for test accuracy versus the contact time, which is obtained from conducting experiments on GBSense  dataset under IID setting using VGG-16, where $\Delta t$ denotes the contact time between GS and satellites.}}
    \label{fig:uplink_bottleneck}
    \vspace{-2ex}
\end{figure}

The impact of resource heterogeneity on satellite-based FL is shown in Fig.~\ref{fig:hetero_resource}. Here, we define the under-training satellite rate as the percentage of under-training satellites (i.e., satellites that cannot participate in model training). From Fig.~\ref{sfig:hetero_compute} and Fig.~\ref{sfig:hetero_memory}, we realize that the results of FedAvg become worse and worse according to ascending under-training satellite rates. Apparently, FedAvg without heterogeneous on-device resource budgets (i.e., Ideal) aggregates all local models from satellites to achieve the fastest training convergence and highest accuracy, but since FedAvg is a static FL framework, heterogeneous on-device resource budgets heavily affect its performance. Therefore, we cannot directly deploy current FL frameworks to LEO satellites. Except for heterogeneous on-device resources, we will meet the uplink communication challenge in the following section.


\vspace{-0.2cm}
\subsection{Bottleneck of Uplink Communication} \label{ssec:uplink_bottlenet}

We leverage a real-world commercial LEO satellite communication system, Starlink to measure and study its downlink and uplink rates. Fig.~\ref{sfig:exStarlink} and Fig.~\ref{sfig:gsStarlink} show  the experimental setup of Starlink and its GS, respectively. We use a well-known network measurement tool, \textit{Iperf}~\cite{tirumala1999iperf} to test and collect downlink and uplink rates, {and their cumulative distribution functions (CDF) are illustrated in Fig.~\ref{sfig:data_rate}} that apparently, the average downlink rate is close to 100\!~Mbps, and larger than the average uplink rate with 12\!~Mbps. In LEO satellite FL, the downlink rate is enough for model aggregation, but the limited uplink rate becomes a major bottleneck for model distribution. For example, the size of a VGG-16 is 528\!~MB, and hence, GS spends about 5.87 minutes to distribute a single model. The time cost of distribution is too long to deliver models to all satellites in each satellite-GS contact time, and thus fewer satellites participate in training their models. \rev{Although specialized GS~(with multi-million US dollars cost~\cite{Devaraj2019PlanetHS,telesat}) offers Gbps- and hundreds of Mbps-level rates respectively for downlink and uplink, multiple tenants need to share these rates,
resulting in very limited per-tenant rates and thus severely restricting their smashed data exchanges. Moreover, recent work~{\cite{vasisht2021l2d2}} also utilizes distributed GSs of low-cost commodity hardware to receive satellite downlink signals. Therefore, results produced by our commercial GS
are representative.}

\begin{figure}[t!]
\centering
\includegraphics[width=8.5 cm]{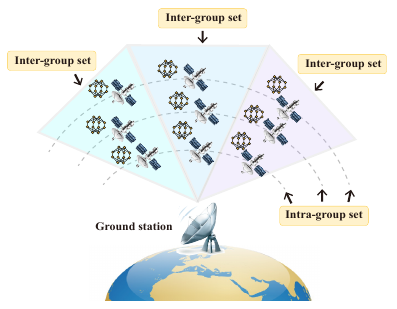}
\vspace{-0cm}
\caption{ The illustration of the intra-group and inter-group sets. The satellites in the triangles of different colors belong to distinct inter-group sets, while satellites within each orbit constitute an intra-group set.}
\label{fig:intra_inter_set}
\end{figure}

We also evaluate local models updating failure at satellite networks. According to our Starlink measurement traces, we use Linux traffic control, tc~\cite{hubert2002linux} to emulate satellite networks condition. We set contact time $\Delta t$ ranging from 5 to 15 minutes~\cite{duggen2007subduction}, and packet loss rate at 100\% when out of contact time immediately via tc. In this experiment, we deploy three satellites\footnote{{In practical deployment, the number of satellites involved during the contact time far exceeds three~\cite{ahmmed2022digital}, resulting in lower transmission rate, thus posing greater challenges for uplink model transmission. Therefore, despite larger GS being equipped with a large antenna with higher transmit power, the uplink model transmission problem is still challenging.}}, and the result of that experiment is shown in Fig.~\ref{sfig:update_fail}. The ideal case assuming no uplink rate limitation has the best performance, but the other two cases lose at least one satellite to participate in training leading to worse performance. The reason is that in the FL model aggregation phase, more satellites can integrate more data into model training~\cite{chen2022federated}. In short, uplink communication of LEO satellites heavily affects FL training performance in practice.


\begin{figure}[t]
\setlength\abovecaptionskip{3pt}
\centering
\subfigure[Intra-group staleness.]{
    \includegraphics[height=2.3cm,width=4cm]{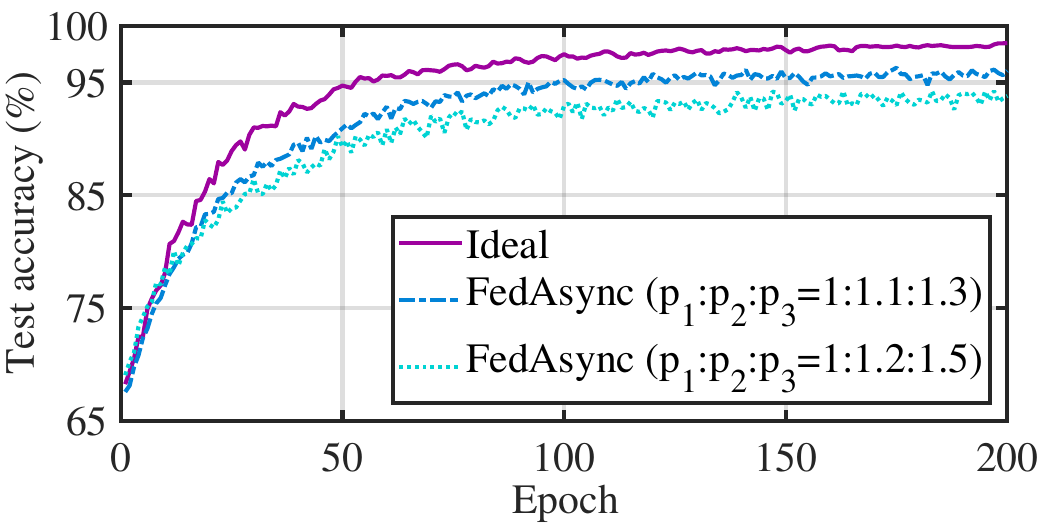}
    \label{sfig:staleness1}
}
\subfigure[Inter-group staleness.]{
    \includegraphics[height=2.3cm,width=4cm]{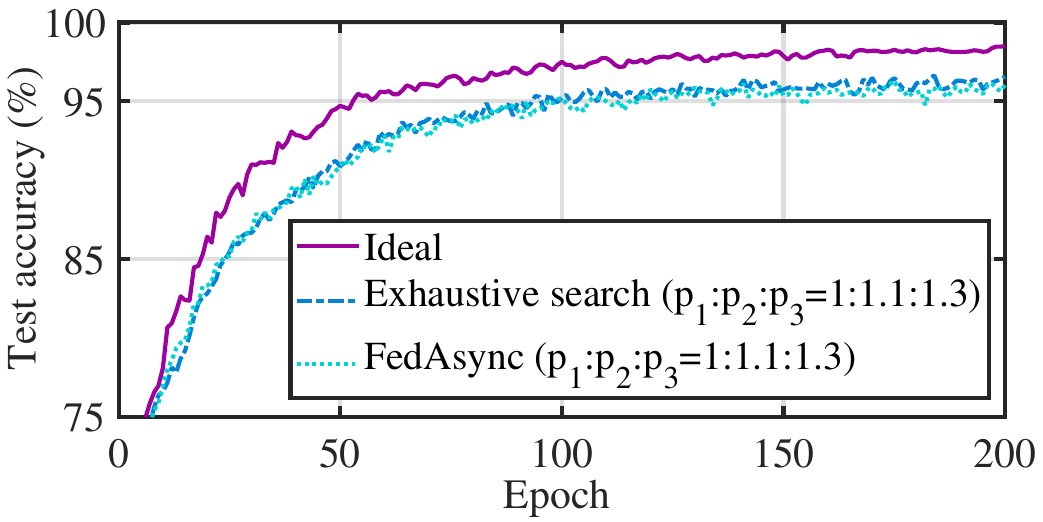}
    \label{sfig:staleness2}
}
    \caption{The impact of intra-group and inter-group staleness on FL. Fig.~\ref{sfig:staleness1} and Fig.~\ref{sfig:staleness2} show the performance for test accuracy versus different intra-group satellite orbital period ratios and inter-group model aggregation schemes, respectively.}
    \label{fig:staleness}
\end{figure}

\vspace{-0.2cm}
\subsection{Staleness in LEO Satellite Networks} \label{ssec:staleness_prob}




Due to intermittent connectivity between the  satellites and a GS, satellites orbiting around the Earth can be divided into two kinds of sets that are inter-group set and intra-group set. Inter-group sets represent all visible satellites of a GS in different contact time, and intra-group sets mean satellites on different orbits during each contact time, illustrated in  Fig.~\ref{fig:intra_inter_set}.  For example, if the contact time between LEO satellites and a GS is 9 minutes and the maximum orbital period is 90 minutes, in this case, the satellites are divided into 10 inter-group sets, and each inter-group set contains multiple intra-group sets.
FL model staleness occurs in both inter-group and intra-group sets. The inter-group model staleness arises from discrepancies in the model version caused by intermittent connectivity, while the intra-group staleness stems from the imbalanced participation of different orbit satellites in FL model training.

We first  study its staleness problem caused by intra-group sets, which are not considered by most of the current FL framework for satellites network~\cite{wu2023fedgsm,elmahallawy2022asyncfleo,razmi2022scheduling}.  For an arbitrary inter-group set, since each intra-group set has its own orbits and moving speed, a GS establishes connections with different intra-group sets to aggregate their local models, but intra-group sets with farther orbits bring higher staleness for the global model. We establish an experiment to motivate that problem happened in intra-group sets. 

In our experimental setup, we set three intra-group sets with orbital period ratios 1:1.1:1.3 and 1:1.2:1.5. {For convenience, we also set three satellites in each intra-group set. We fix  FedAsync scheme~\cite{xie2019asynchronous} for inter-group staleness to update the global model whenever the local models at different intra-group sets are available, and results are shown in Fig.~\ref{sfig:staleness1}. }
The third intra-group set uploads its local models of satellites may bring negative gradients to impact the global model even with staleness compensation, since their local models are outdated compared to those of the first and second intra-group sets.

\begin{figure}[t!]
\centering
\includegraphics[height=6.65cm, width=8.5 cm]{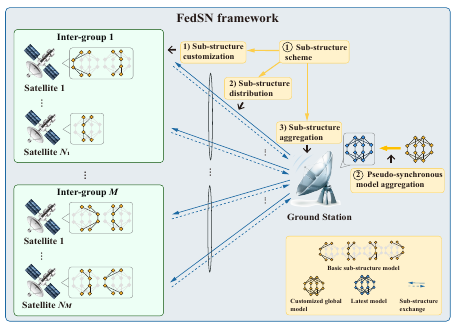}
\vspace{-0cm}
\caption{The system overview of \name framework.}
\label{fig:FedSat_overview}
\end{figure}


For staleness problems caused by inter-group sets, several previous works have been studied~\cite{wu2023fedgsm,elmahallawy2022asyncfleo,razmi2022scheduling,so2022fedspace}, but all of them only consider  time-dependent compensation in their FL frameworks. We keep the above experimental setting and set 10 inter-group sets in this experiment. While we aggregate local models with time-dependent compensation, its performance is worse than the ideal approach assuming no inter-group sets shown in Fig.~\ref{sfig:staleness2}.  However, we perform 
an exhaustive search to figure out the best set~(9 inter-groups in our experiment) to aggregate their local models, and obtain performance improvement compared with only the time-dependent compensation approach illustrated in Fig.~\ref{sfig:staleness2}. Therefore, we need to reconsider the discrepancies in satellites' local models for FL aggregation.   


\section{Framework Design} \label{sec:system_design}

\subsection{Overview of \name}

Motivated by Section~\ref{sec:background_motivation}, we propose and design a framework, named \name to tackle the above challenges. \name comprises two main components: sub-structure scheme and pseudo-synchronous model aggregation. The sub-structure scheme includes i) sub-structure customization, ii) sub-structure distribution, and iii) sub-structure aggregation. Fig.~\ref{fig:FedSat_overview} illustrates that for the sub-structure scheme, we only consider intra-group sets. We employ a sub-structure scheme to handle all visible satellites' heterogeneous on-device resources~(computing, memory, and uplink bandwidth). The GS leverages sub-structure customization to customize basic sub-structure models and determine the number of these models for each satellite  under its on-device resources budget. Then, the sub-structure distribution is used to select the best combination of basic sub-structure models for each visible satellite and transmit all combinations to all visible satellites. After training, the sub-structure aggregation is utilized to aggregate basic sub-structure models for subsequent model assembling. For pseudo-synchronous model aggregation, we employ pseudo-synchronous model aggregation to aggregate all customized global models from different inter-groups into a global model.

\vspace{-0.3cm}
\subsection{FL Framework over LEO Satellite Networks}
In this section, we elaborate on the FL framework over LEO satellite networks to provide a theoretical foundation for the following sub-structure scheme and pseudo-synchronous model aggregation.
As illustrated in Fig.\ref{fig:Fed_spa_scenario}, we consider a typical scenario of \name over the  satellite networks, which consists of two primary components:
\begin{itemize}
\item \textbf{LEO satellites: }We consider that all satellites possess specific budgets (i.e., on-device computing and memory resources) for model training. The set of satellites involved in model training is denoted by $\mathcal{N} = \left\{ {1,2,..., N} \right\}$, where $N$ is the number of satellites.  The collections of on-device computing and memory resources are represented as $\mathcal{C} = \left\{ {c_1,c_2,...,c_N} \right\}$ and $\mathcal{S} = \left\{ {s_1,s_2,...,s_N} \right\}$, where ${c_i} \in \left[ {0,1} \right]$ and ${s_i} \in \left[ {0,1} \right]$ denote available resources for the $i$-th satellite (i.e., a fraction of  global model corresponding to computing workload and memory space that $i$-th satellite can afford.). Similarly, the impact of uplink rates on model training is denoted by $\mathcal{U} = \left\{ {u_1,u_2,...,u_N} \right\}$, where ${u_i} \in \left[ {0,1} \right]$ denotes a fraction of global model that the GS can upload to $i$-th satellite during a period of contact time. The local dataset $\mathcal{D}_i$ 
residing on the $i$-th satellite is denoted by ${\mathcal{D}_i} = \left\{ {{{\bf{x}}_{i,k}},{y_{i,k}}} \right\}_{k = 1}^{{D_i}}$, where ${\bf{x}}_{i,k}$ and ${y}_{i,k}$ represent the $k$-th input data in the local dataset $\mathcal{D}_i$ and its corresponding label. Thus, the total dataset 
is ${\cal D} =  \cup _{i = 1}^N{{\cal D}_i}$.

\item \textbf{Ground station: }{The ground station serves as a parameter server with powerful computing capability, and thus executes the sub-structure scheme and pseudo-synchronous model aggregation for \name.} Concurrently, the ground station is also obligated to collect parameters of each visible satellite including computing and memory resources, as well as channel state information for our framework \name operation.


\end{itemize}

The global model, assumed to be a convolutional neural networks (CNNs) that  are widely used for vision tasks, is denoted by ${{\bf{W}}} \in {\mathbb{R}^{b}}$, where $b$ is dimension of model parameters. The predicted value derived from the input data ${{\bf{x}}_{i,k}}$ is represented as ${{{\hat y}}_{i,k}} = f\left( {{{\bf{x}}_{i,k}};{{\bf{W}}}} \right)\in {\mathbb{R}^{q}}$, where $f\left( {{\bf{x}};{\bf{w}}} \right)$ maps the relationship between the input data $\bf{x}$ (we abuse $\bf{x}$ here) and the predicted value given model parameter ${\bf{w}}$. Therefore, the local loss function for the $i$-th satellite is denoted by ${L_i}\left( {\bf{W}} \right) = \frac{1}{{ {{ |\mathcal{D}_i| }} }}\sum\limits_{k = 1}^{ {|\mathcal{D}_i|} } {{L_{i,k}}\left( {{{\bf{x}}_{i,k}},{y_{i,k}}};{\bf{W}} \right)} $, where ${{L_{i,k}}\left( {{{\bf{x}}_{i,k}},{y_{i,k}}};{\bf{W}} \right)}$ represents the sample-wise loss function for the $k$-th data sample in the local dataset ${\mathcal{D}_i}$. The global loss function is the weighted average of the local loss functions, where the weights are proportional to the local dataset size. The objective of FL is to find the optimal model parameter ${{\bf{W}}^{\bf{*}}}$ in the following optimization problem:
\begin{align}\label{eq:global_loss_minimization}
\mathop {\min }\limits_{\bf{W}} L\left( {\bf{W}} \right) = \mathop {\min }\limits_{\bf{W}} \sum\limits_{i = 1}^N {\frac{{{D_i}}}{D}} {L_i}({\bf{W}}).
\end{align}
To solve Eqn.~\eqref{eq:global_loss_minimization}, conventional FL (e.g., FedAvg~\cite{mcmahan2017communication}) employs synchronous model aggregation to seek the optimal model parameter. However, as we stated and observed in Sec.~\ref{sec:background_motivation},  intermittent connectivity between the LEO satellites and GS renders direct implementation of FedAvg impractical. Furthermore, in conventional FL, the dropout of satellites with limited available on-device resources (satellites with ${c_i} < 1$ or ${s_i} < 1$ or ${u_i} < 1$), and the imbalance participation of satellites caused by various orbital periods  lead to severe model bias, and hence, a degradation in training performance. To overcome the aforementioned challenges introduced in Sec.~\ref{sec:background_motivation}, we give design details about a new-brand FL framework over LEO satellite networks, incorporating intra-group sub-structure scheme and inter-group pseudo-synchronous model aggregation in the following sections.

\vspace{-0.15cm}
\subsection{Satellite Characterization}
Recalling Fig.~\ref{fig:intra_inter_set}, LEO satellites in various orbits around the Earth are divided into $K$ inter-group sets. For the $k$-th satellite set, satellites participated in model training
is denoted by $\mathcal{N}^k = \left\{ {1,2,...,N_k} \right\} \subseteq {\cal N}$, where $N_k$ is the number of satellites in the $k$-th satellite set. The local dataset residing in the $i$-th satellite in the $k$-th inter-group set is represented as ${\mathcal{D}^k_i}$, and thus the total dataset is denoted by ${{\cal D}^k} =  \cup _{i = 1}^{N_k}{{\cal D}_i^k}$. The collections of computing and memory resources of satellites in $k$-th 
inter-group set are represented as ${{\cal C}^k} = \left\{ {c_1^k,c_2^k,...,c_{{N_k}}^k} \right\} \subseteq {\cal C}$ and $\mathcal{S}^k = \left\{ {s^k_1,s^k_2,...,s^k_{N_k}} \right\} \subseteq {\cal S}$, where ${c^k_i} \in \left[ {0,1} \right]$ and ${s^k_i} \in \left[ {0,1} \right]$ denote the available computing and memory resources for the $i$-th satellite in the $k$-th inter-group set. {The orbital period of the $j$-th satellite orbits ($j$-th intra-group set) is represented as $p_j$.} We calculate ${c^k_i}$ and ${s^k_i}$ as
\begin{align}\label{eq:c^k_i}
c_i^k = \min \left\{ {1,\frac{{\gamma^k_i {p_j}}}{{\eta \theta }}} \right\},\quad \quad i = 1,2,...,{N_k},
\end{align}
and
\begin{align}\label{eq:s^k_i}
s_i^k = \min \left\{ {1,\frac{{{\varpi^k_i} '}}{\varpi }} \right\},\quad \quad i = 1,2,...,{N_k},
\end{align}
where $j = \left\{ {m\mid i \in {{\mathcal{P}}_m}} \right\}$, ${\mathcal{P}}_m$ represents the set of satellites in the $m$-th orbit (i.e., the $m$-th intra-group set), {$\frac{{\gamma^k_i {p_j}}}{{\eta \theta }}$ is the proportion of global model training completed by $i$-th satellite in the $k$-th inter-group in an orbital period, ranging from 0 to 1,} $\gamma^k_i$ denotes the computing resources allocated to FL model training for the $i$-th satellite in the $k$-th inter-group set (namely, the number of floating point operations (FLOPs) per second), $\eta$ and $\theta$ are local update epoch (training round) and the computing workload (in FLOPs) of global model training for a mini-batch data samples, ${\varpi^k_i} '$ and $\varpi$ represent the available memory resources for the $i$-th satellite in the $k$-th inter-group set and the data size (in bits) of the global model, respectively.

To characterize the impact of uplink quality on satellites' available link budget, we define ${u^k_i} \in \left[ {0,1} \right]$ to represent the fraction of hidden layer channels for sub-structure models that the GS can upload to the $i$-th satellite  in the $k$-th inter-group set during a period of contact time. We formulate ${u^k_i}$ as
\begin{align}\label{uplink_fraction}
u_i^k = \min \left\{ {1,\frac{{r_i^k\Delta t}}{\phi }} \right\},\quad \quad i = 1,2,...,{N_k},
\end{align}
where $r^k_i= \min \left\{ {r^{k}_{u,i}, r^{k}_{d,i}} \right\}$, $r^{k}_{u,i}$ and $r^{k}_{d,i}$ are the average uplink and downlink rates between GS and the $i$-th satellite in the $k$-th inter-group set, and $\Delta t$ and $\phi$ represent a period of contact time and data size (in bits) of a global model, separately. For our design, jointly considering satellites' heterogeneous computing and memory resources as well as uplink rate, the available budget for the $i$-th satellite in the $k$-th inter-group set is expressed as 
\begin{align}\label{available_buget}
b_i^k = \min \left\{ {c_i^k,s_i^k,u_i^k} \right\},\quad \quad i = 1,2,...,{N_k}.
\end{align}
Before the model training begins, a GS gathers an available budget for each satellite, and according to that information, \name can execute the following intra-group sub-structure scheme. 

\begin{figure}[t!]
\centering
\includegraphics[height=6.65cm, width=8.5 cm]{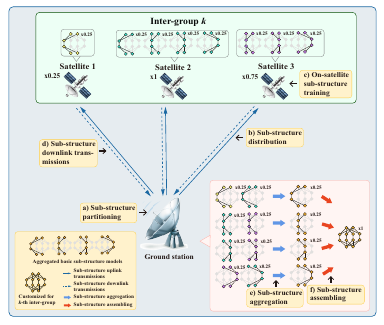}
\vspace{-0cm}
\caption{ An example of intra-group training procedure of \name for the $k$-th inter-group set, where three satellites in the set participate in model training with the available budget ${{\cal B}^k} = \left\{ {0.25,1,0.75} \right\} $.
}
\label{fig:Sub_structure_scheme}
\end{figure}

\vspace{-0.2cm}
\subsection{Intra-group Sub-structure Scheme Workflow}\label{sec:sub_scheme}
 
In this section, we introduce our intra-group sub-structure scheme that comprises sub-structure customization, distribution, and aggregation methods. The fundamental idea underlying the proposed scheme is to split complete knowledge into multiple smaller shards and then customize specific shards via flexible assembling. Without loss of generality, we focus on the $k$-th inter-group set for analysis. As depicted in Fig.~\ref{fig:Sub_structure_scheme}, at an arbitrary training round\footnote{Here, the duration of one training round is defined to coincide with the maximum orbital period. Thus, the model training procedure within the maximum orbital period is called one training round.} $t \in \mathcal{T} = \left\{ {1,2,..., T} \right\}$, the intra-group training procedure consists of the following five steps. For simplicity, the index $t$ of the training round number is omitted.

a) \textit{Sub-structure Partitioning}: At this step, the global model is first split into several basic sub-structure models ${{\mathcal W}^k} = \left\{ {{{\bf{w}}^k_1},{{\bf{w}}^k_2},...,{{\bf{w}}^k_L}} \right\} $ by channel-wise partitioning based on the minimum constrained budget in the $k$-th inter-group set, where ${\bf{w}}^k_l$ and $L$ represent the $l$-th basic sub-structure model and the number of basic sub-structure models, respectively. Then, a GS determines the number of basic sub-structure models that each satellite can afford according to the heterogeneous satellites' available budgets, which will be elaborated in Sec.~\ref{sec:sub_cus}.

b) \textit{Sub-structure Distribution}: Afterward, the GS extracts combinations of basic sub-structure models ${\cal W}_1^k,{\cal W}_2^k,..., {\cal W}_{{N_k}}^k \subseteq {\cal W}^k$ for each satellite and then distributes selected combination to the corresponding satellite via the uplink. We will discuss how to extract subsets from ${\cal W}^k$ in Sec.~\ref{sec:sub_dis}.

c) \textit{On-satellite Sub-structure Training}: After receiving the combination of basic sub-structure models from the GS, each satellite utilizes the locally residing data to train its corresponding basic sub-structure models in parallel.

d) \textit{Sub-structure Downlink Transmissions}: Each satellite sends its combination of basic sub-structure models to the GS within the contact time after local model training.  


e) \textit{Sub-structure Aggregation}: At this step, the GS aggregates the same basic sub-structure models collected from the different satellites into one for model assembling. The sub-structure aggregation method tailored for intra-group model staleness caused by orbital period difference is illustrated in Sec.~\ref{sec:sub_agg}.

f) \textit{Sub-structure Assembling}: After aggregating the same basic sub-structure models, the GS proceeds to assemble a customized global model based on those aggregated basic sub-structure models, which is also demonstrated in Sec.~\ref{sec:sub_cus}.

The workflow of intra-group training of \name has been outlined above. The tailored sub-structure scheme incorporates sub-structure customization, distribution, and aggregation methods in the aforementioned stages a), d), and e). In the following sections, we will discuss their design challenges and provide corresponding design details for each one.

\begin{figure}[t]
\vspace{-.5ex}
\setlength\abovecaptionskip{3pt}
\centering
\subfigure[Model width.]{
    \includegraphics[height=2.3cm,width=4cm]{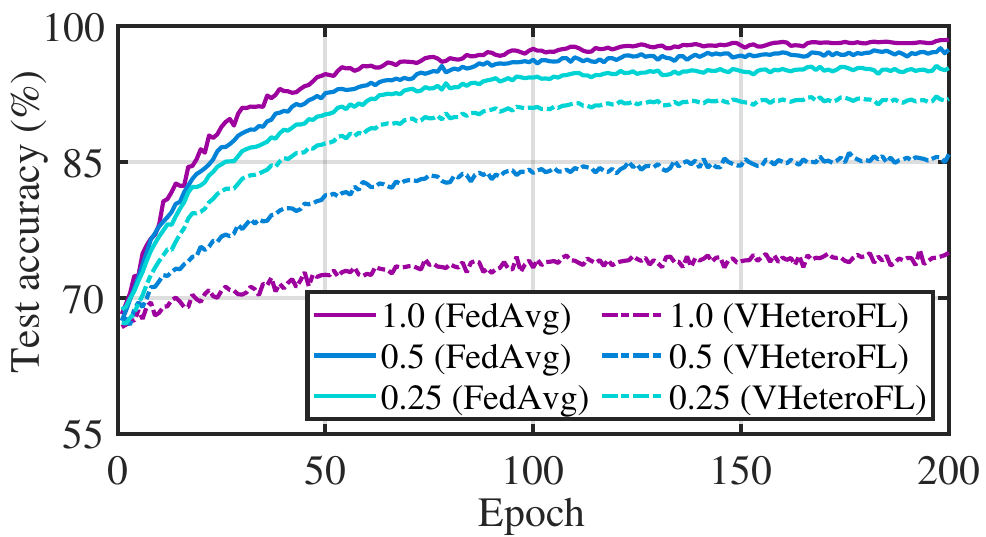}
    \label{fig:sub_extraction}
}
\subfigure[Distribution scheme.]{
    \includegraphics[height=2.3cm,width=4cm]{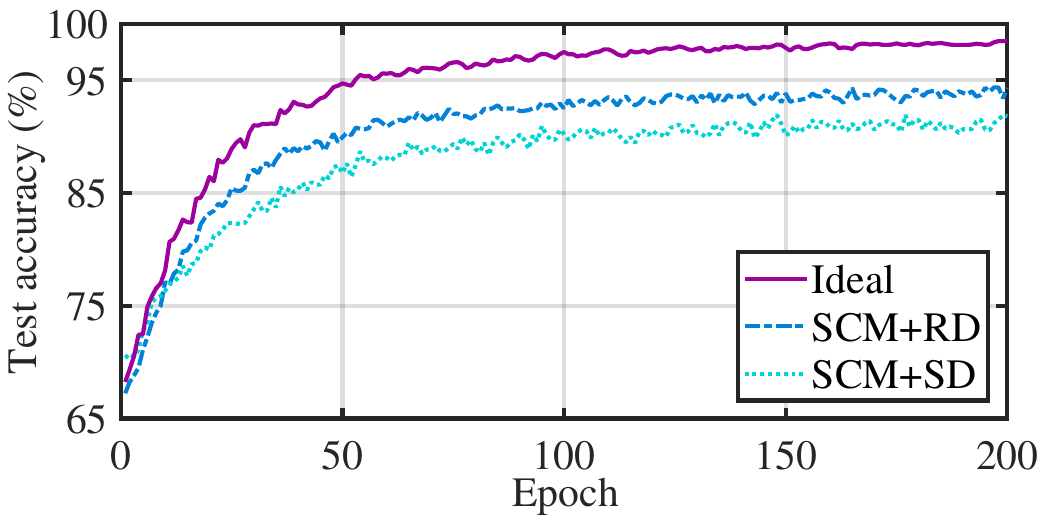}
    \label{fig:sub_distribution}
}

    \caption{Convergence of models with different width (a) and varying distribution schemes (b) on FL for GBsense dataset under IID setting.}
    \label{fig:sub_struc_width_distribution}
    
    \vspace{-2ex}
\end{figure}

\vspace{-0.3cm}
\subsection{Intra-group Sub-structure Scheme (SS)}\label{sec:sub_scheme}

\subsubsection{Sub-structure Customization Method (SCM)}\label{sec:sub_cus}



Recalling that Sec.~\ref{ssec:hrtero_compute_memory} and Sec.~\ref{ssec:uplink_bottlenet}, to combat the challenges caused by heterogeneous available budgets, we need to devise flexible and scalable model customization method. Before customizing models, it is essential to evaluate their complexities.
The predominant metrics for quantifying model complexity are the width~\cite{zagoruyko2016wide} or depth~\cite{tan2019efficientnet}~(i.e., the number of channels or layers in a model). Even though there is a learning framework, called \textit{split learning} that splits a global model into client-side and server-side models with different depths, and trains them in both two sides~\cite{lin2023split,lin2023pushing}, clients must share features of their client-side models with the server during a training phase. Considering LEO satellite networks, limited contact time is too short to train a good and robust model via split learning. Consequently, 
in this paper, we focus on the model width for our sub-structure model design.

{As shown in Fig.~\ref{fig:Sub_structure_scheme}, consistent with~\cite{zagoruyko2016wide}, we use the number of channels in the hidden layer as the model width to construct varying basic sub-structure models, but still keep their input and output layers unchanged.}
Flexible control of the model width enables diverse sub-structure complexities. This is because {different sub-structure models}
have various numbers of parameters, leading to {satisfying heterogeneous available budgets for satellites.} {However, to design the sub-structure customization method is very challenging for satellites, and thus, we demonstrate that challenge by studying how diverse model widths affect training performance.
} 

{
We first study \textit{budget-unconstrained} FL over LEO satellite networks. Local models with different widths are trained individually by utilizing FedAvg, but their maximum widths are limited by $\times$1, $\times$0.5, and $\times$0.25. Their training performance is illustrated in Fig.~\ref{fig:sub_extraction} with solid lines.}
It is clear that wider models exhibit faster convergence speed and better generalization than the {narrow models}. 
{Therefore, those experiments indicate that we need to customize wider models, but not narrower ones.}

{
We also study \textit{budget-constrained} FL training performance via employed HeteroFL~\cite{diaoheterofl}. HeteroFL is 
the status quo FL framework for heterogeneous clients with different computing capabilities, but it only assumes the local model of each satellite has a pre-defined fixed width. For example, if the $i$-th satellite has the specific computing resource $c_i^{k} = 0.75$, HeteroFL can only assign the corresponding fixed width $\times$0.75 model, but cannot assign the other fixed width models. Hereby, the $\times$0.75 model cannot be trained using the other datasets on their satellites. Consequently, HeteroFL fails to fully utilize the dispersed data across satellites during its training phase.
We modify HeteroFL to a variant of HeteroFL, named VHeteroFL, and evaluate  its training performance, where each satellite can train as many affordable models as possible. It implies that while the satellite with $c_i^{k} = 0.75$, it can train $\{ \times0.25, \times0.5, \times0.75 \}$ models.}
To simplify the analysis, 
the available budgets for satellites are 
{configured} to follow a discrete uniform distribution, i.e., $P(b_i=0.25, 0.5, 0.75, 1)= {0.25}$. 
{
The results with dashed lines are shown in Fig.~\ref{fig:sub_extraction}. 
}
It is clear that the wider models not only exhibit lower test accuracy than their FedAvg counterparts but even narrower models. This is because data from only 1/4 of the satellites is available for training the $\times$1 models, while data information from 3/4 of the satellites is integrated into the $\times$0.25 models. The increased data involvement empowers the $\times$0.25 models with better generalization than the $\times$1 models, indicating that customizing the widest model may not be a wise choice.

Given the above two observations, we find that the sub-structure customization needs to be well-designed, otherwise, it may lead to deterioration rather than improvement in training performance. In view of this, we propose a flexible sub-structure customization method.
{We first determine basic sub-structure models based on the minimum budget constraint within satellite groups, ensuring that every satellite is capable of training these models.
Then, we split a global model into multiple basic sub-structure models via channel-wise partitioning and customize different combinations of those models according to satellites' available budgets ${{\cal B}^k} = \left\{ {b_1^k,b_2^k,...,b_{{N_k}}^k} \right\}$.} 
This eliminates occurrences of under training for basic sub-structure models, thereby effectively improving the convergence speed and test accuracy of model training.
{We use an example to illustrate how \name determines the number of basic sub-structure models in Fig.~\ref{fig:Sub_structure_scheme}.} 
\begin{figure}[t]
	\centering
	\includegraphics[width=8cm]{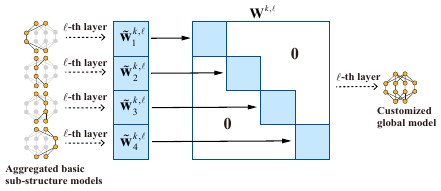}
	\vspace{-0cm}
	\caption{{An example of model assembling for $k$-th inter-group set, where the $\ell$-th layer weight matrices of the $l$-th aggregated basic sub-structure models are ${\bf{\tilde w}}_l^{k,\ell }$, and the $\ell$-th layer weight matrix of the customized global model is ${\bf{W}}^{k,\ell}$.} 
	}
	\label{fig:sub_assemble}
\end{figure}

The global model is split into four $\times$0.25 basic sub-structure models, and there are three satellites in the $k$-th inter-group set with {${{\cal B}^k} = \left\{ {0.25,0.75,0.75} \right\} $.} In this case, the GS 
{determines that the number of basic sub-structure models} are $1$, $3$, and $3$ , for {those satellites,} respectively.

\begin{figure*}[t]
	\centering
	\includegraphics[width=17.8 cm]{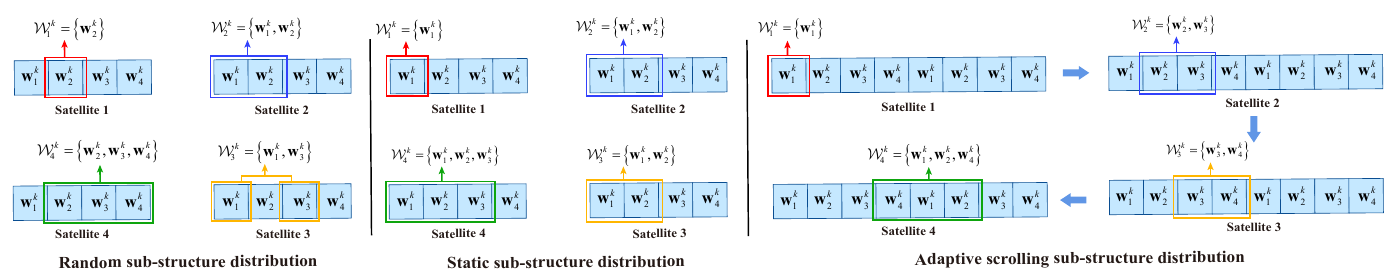}
	\vspace{-0cm}
	\caption{An example of how sub-structure combinations are extracted by random sub-structure distribution (left), static sub-structure distribution (middle), and the proposed adaptive scrolling sub-structure distribution (right), where ${\mathcal{W}^k_i}$ denotes the selected sub-structure combination for $i$-th satellite in the $k$-th inter-group set, the global model is partitioned into four $\times0.25$ basic sub-structure models (i.e., $L=4$) and the available budgets are ${{\cal B}^k} = \left\{ {0.25, 0.5, 0.5, 0.75} \right\} $ .}
	
	\label{fig:Sub_distribution_method}
\end{figure*}

{After basic sub-structure models partitioning, we need to assemble them into a customized global model\footnote{{The sub-structure customization method only contains partitioning and assembling. Although there are sub-structure distribution and aggregation steps between partitioning and assembling, they are leveraged to improve overall training performance in LEO satellite networks.}}. Since each satellite trains its combination of basic sub-structure models under its available budget, the budget-constrained FL problem is relaxed to the budget-unconstrained FL problem. Recalling that Fig.~\ref{fig:sub_extraction}, wide models show faster convergence speed and better generalization than the narrow model for budget-unconstrained FL. Consequently, our key idea is to assemble the global model as widely as possible.} 

{Motivated by dropout technique~\cite{krizhevsky2012imagenet} that zeroes out the activation of some weights to prevent overfitting and sometimes even improves training efficiency, our sub-structure scheme sacrifices some weights in exchange for the flexibility of training heterogeneous models. Moreover, since the sub-structure models are determined based on the minimum budget of the satellites and each substructure can extract the data information from all satellites, zeroing out some weights may not lead to significant performance deterioration.}

As demonstrated in Fig.~\ref{fig:sub_assemble}, the global model is partitioned into four $\times$0.25 sub-structures based on hidden layer channels. {For clarity, we reshape the weights of the $\ell$-th layer of global model into the weight matrix ${\bf{W}}^{\ell} \in {\mathbb{R}^{p \times q}}$, where $p$ and $q$ represent the number of input and output channels, respectively.}  {To reconstruct the global model, the weight matrices of aggregated basic sub-structures are concatenated along the diagonal shown in Fig.~\ref{fig:sub_assemble} into a matrix with the same size as the global model's weight matrix. }During the matrix concatenation process, the non-trainable parameters (grey areas) are set to $0$ to avoid interference between different basic sub-structures. Hence, for any given $\ell$-layer, the weight matrix of the assembled global model is represented as
\begin{align}\label{assembled_weight}
	{{\bf{W}}^{k,\ell} } \!= \!\left\{ {\begin{array}{*{20}{c}}
			{\!\!{\bf{\tilde w}}_1^{k,\ell} }&{}&{}&\bf{\textit{O}}\\
			{}&{{\bf{\tilde w}}_2^{k,\ell} }&{}&{}\\
			{}&{}& \ddots &{}\\
			\bf{\textit{O}}&{}&{}&{{\bf{\tilde w}}_L^{k,\ell} }\!\!
	\end{array}} \right\}
\end{align}
where ${\bf{\tilde w}}_l^{k,\ell }$ is the $\ell$-th layer weight matrices of the $l$-th aggregated basic sub-structure models for the the $k$-th inter-group set.

\subsubsection{Sub-structure Distribution Method (SDM)}\label{sec:sub_dis}

When determining the number of basic sub-structure {models for diverse satellites, 
for their combinations, the basic sub-structure models must be selected, and distributed. We discuss how to figure out appropriate combinations for each satellite in this section.}

{To study the importance of distribution methods for our \name,}
we set up the sub-structure customization method in Sec.~\ref{sec:sub_cus} with $L=4$, {and} deploy two distinct sub-structure distribution approaches: random distribution (RD) and static distribution (SD) {in our experiment}. {For the RD and the SD, sub-structure combinations are extracted from a global model in random and fixed manners, respectively. As illustrated in Fig.~\ref{fig:Sub_distribution_method}, the RD randomly picks up basic sub-structure models to form the combination for each satellite, but the SD keeps fixed combinations.
}
Fig.~\ref{fig:sub_distribution} shows the significant performance gap {among RD}, {SD}, and the ideal case, where the ideal performance is achieved by deploying a global model on each satellite for local model training. There are two {insights} about these results:
\begin{itemize}
\item For {the RD}, it randomly extracts different combinations in consecutive training rounds, {resulting in uneven training on basic sub-structure models. That uneven training brings failures to balance update frequencies of different parts of the global model, inevitably leading to slower convergence speed and lower test accuracy.
}



\item For {the SD}, performance degradation comes from two major inherent drawbacks: First, {as illustrated in Fig.~\ref{fig:Sub_distribution_method}, in most cases, satellites cannot train all basic sub-structure models, except that some of them have enough available budgets for an entire global model. Second, due to fixed combinations, any basic sub-structure model is not fully trained on dispersed data across all four satellites. 
}

\end{itemize}

According to the above {insights}, it is crucial to carefully design the sub-structure distribution method for improving training performance. We propose an adaptive scrolling sub-structure distribution method that adheres to the following two design principles: First, 
different basic sub-structure models should be trained as equally as possible. Second, the dataset dispersed across distinct satellites should be {trained on each basic sub-structure model,}
regardless of the satellites' available budgets. {
In each round, we utilize an adaptive scrolling window in each satellite based on its available budget to select sequential basic sub-structure models, but for different satellites, we need to ensure that the window does not start from the same position in the sequence of basic-sub-structure models.  {This dynamic scrolling of substructure combinations prevents fixed combinations from training exclusively on a single satellite, thereby guaranteeing the evenness of model training.}
}


\begin{figure*}[t!]
\centering
\includegraphics[width=14.5 cm]{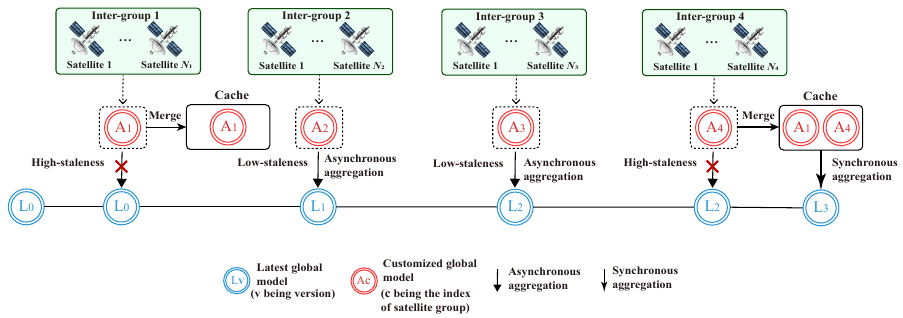}
\vspace{-0cm}
\caption{An example of the workflow of the proposed pseudo-synchronous aggregation strategy for one training round, where satellites orbiting around the Earth are divided into $4$ groups (i.e., $K = 4$).}
\label{fig:asyn_agg_illu}
\end{figure*}

Taking Fig.~\ref{fig:Sub_distribution_method} as an example, considering the available budget ${{\cal B}^k} = \left\{ {0.25, 0.5, 0.5, 0.75} \right\} $ for the $k$-th satellite group, the corresponding cardinalities of the {combination} are $1$, $2$, $2$, and $3$, respectively. For the first satellite, we select the {basic sub-structure model ${{{\bf{w}}^k_1}}$} using a window of length $1$. Then, the scrolling window {moves forward one step}, and its length is adjusted to $2$ based on the available budget of the second satellite, resulting in the combination $\left\{ {{{\bf{w}}^k_2},{{\bf{w}}^k_3}} \right\}$. Similarly, the combinations $\left\{ {{{\bf{w}}^k_3},{{\bf{w}}^k_4}} \right\}$ and $\left\{ {{{\bf{w}}^k_1},{{\bf{w}}^k_2}},{{\bf{w}}^k_4} \right\}$ are {selected} to the third and fourth satellites, respectively.

In a nutshell, on the one hand, the scrolling window mechanism ensures that {all basic sub-structure models are trained approximately equally.}
Additionally, the adaptive window size allows real-time adjustment based on the available budgets of different satellites. On the other hand, {this method fully train data dispersed across distinct satellites. In the following section, we will describe the sub-structure aggregation method.}


\subsubsection{Sub-structure Aggregation Method (SAM)}\label{sec:sub_agg}

The same {basic sub-structure models} from different satellites need to be aggregated for model assembling. However, as described in Sec.~\ref{sec:background_motivation}, the varying orbital periods of satellites in different orbits within the satellite group lead to an imbalance in satellite participation in model training, thereby lowering the effectiveness of model training. 
{To compensate the model bias in FL model training, it is essential to develop an appropriate sub-structure aggregation method.}


Since the sub-structure aggregation within a group is synchronous, we directly incorporate a corrective term into the aggregation weights to offset model bias caused by orbital period differences. {Satellites with shorter orbital periods can aggregate their updated local models more frequently than those with longer orbital periods, leading to a model bias towards satellites with shorter orbital periods and thus deteriorating the training performance. To address the above issue, the corrective term gives smaller aggregation weights to satellites with shorter orbital periods and larger aggregation weights otherwise, to balance the model aggregation frequency across different satellites. } The orbital period ratio of the $k$-th {inter-group set} is assumed as $p_1:p_2,...,:p_J \, (p_J \ge p_{J - 1},...,p_1)$, where $J$ represents the number of satellite orbits. The aggregated basic sub-structure {models} are given by 

\begin{align}\label{aggregated_weight}
{{{\bf{\tilde w}}}_l} = \sum\limits_{i \in {\cal H}_l^k} {\frac{{{p_j}}}{{{p_J}}}\delta _i^k{\bf{w}}_{l,i}^k} ,\quad l = 1,2,...,L,
\end{align}
{where $\delta _k^l = \frac{{\left| {\mathcal{D}}_i^k \right|}}{{{\left| {\mathcal{D}}^k \right|}}}$ represents the aggregation weights that are proportional to the local dataset sizes}, $j = \left\{ {m\mid i \in {{\mathcal{P}}_m}} \right\}$, ${\cal H}_{l}^{k} =\left \{ i\mid {\bf{w}}^k_{l} \in {\cal W}_{i}^{k}   \right \}$ is the set of satellites with the $l$-th basic sub-structure model (i.e., ${\bf{w}}^k_{l}$), ${\bf{w}}_{l,i}^k$ denotes the $l$-th basic sub-structure model from the $i$-th satellite in the $k$-th inter-group set.


\vspace{-0.2cm}
\subsection{Inter-group Pseudo-synchronous Model Aggregation Strategy (PMAS)}
\vspace{-0.1cm}


Motivated by {Sec.~\ref{sec:background_motivation}}, we propose a pseudo-synchronous model aggregation strategy. The core idea of this strategy is to first asynchronously aggregate {local} models with low staleness to accelerate the convergence speed of model training, and then synchronously aggregate models with higher staleness at the end of {each} training round to enhance generalization. {To this end,} unlike existing approaches {~\cite{wu2023fedgsm,elmahallawy2022asyncfleo,razmi2022scheduling,so2022fedspace}}, {the strategy considers the similarity between model weights to characterize the extent of model staleness.} Meanwhile, our strategy does not consistently perform asynchronous aggregation but adopts a buffer-based method. 
Without loss of generality, we focus on an arbitrary training round $t \in \mathcal{T} = \left\{ {1,2,...,T} \right\}$ consisting of $K$ {inter-group sets}, and for simplicity, the index $t$ of training round number is omitted. The pseudo-synchronous model aggregation strategy includes the following steps:

a) \textit{Weighting Function Design}: During the contact time between the $k$-th {inter-group set} and GS, our strategy evaluates model weights similarity between the {customized} model of that group ${{\bf{W}}^k}$ and the latest model ${{\bf{W}}^*}$ stored in the GS. The cosine distance is utilized to assess discrepancies between model
weights, which can be calculated as:
\begin{align}\label{eq:cos_dis}
{d^k} = 1 - \frac{{{{{\bf{\hat W}}}^k} \!\cdot\!{{{\bf{\hat W}}}^*}}}{{ \Vert {{{{\bf{\hat W}}}^k}} \Vert \Vert {{{{\bf{\hat W}}}^*}} \Vert}},\quad k = 1,2,...,K,
\end{align}
where ${{\bf{\hat W}}^*} = vec\left( {\bf{W}}^* \right)$, ${{{\bf{\hat W}}}^k} = vec\left( {{{\bf{W}}^k}} \right)$, and $vec\left(  \cdot  \right)$ denotes the vectorization operation of a matrix.

According to Eqn.~\eqref{eq:cos_dis}, model staleness function is formulated as
\begin{align}\label{model_stale}
S\left( {d^k,\tau^k } \right) = {d^k}s\left( \tau^k  \right),
\end{align}
where $ s\left( \tau  \right)$ is weighting function\footnote{We can adopt various time-dependent weighting functions available, such as Constant, Polynomial, and Hinge~\cite{xie2019asynchronous}. In practical deployments, the selection of the weighting function can be determined based on the training task's tolerance towards model staleness.}, {${\tau ^k}$ denotes the average time difference characterizing the model staleness for the $k$-th inter-group set, defined as the number of inter-group sets connected by GS between the current and the last model aggregations.}

b) \textit{Low-staleness Asynchronous Model Aggregation}: 
{
For the $k$-th {inter-group set}, the latest model stored in the GS is updated via
\begin{align}\label{eq:asyn_update}
\begin{small}    
{{\bf{W}}^*} \!\!=\! \left\{ {\begin{array}{*{20}{c}}
\!\!\!\!{\left( {1\! - \!S\left( {{d^k},{\tau ^k}} \right)} \right)\!{{\bf{W}}^*}\!\! + \!\alpha S\left( {{d^k},{\tau ^k}} \right)\!{{\bf{W}}^k},S\left( {{d^k},{\tau ^k}} \right) \le {\gamma _{th}},}\\
{\quad \quad \quad \quad \quad \quad {{\bf{W}}^*},\quad \quad \quad \quad \quad \quad \quad \quad \!\!S\left( {{d^k},{\tau ^k}} \right) > {\gamma _{th}},}
\end{array}} \right.
\end{small}
\end{align}
where $\alpha$ is the mixing hyperparameter of asynchronous aggregation. The threshold $\gamma _{th}$ for model staleness is determined {empirically} according to the training task. 
GS switches the appropriate strategy for the customized model of any inter-group set according to Eqn.~\eqref{eq:asyn_update}.}
Specifically, when the value of the staleness function for {customized} model is below {the threshold $\gamma _{th}$}, it is classified as a low-staleness model and asynchronously aggregated into the latest model, otherwise it is deemed as a high-staleness model and stored in the cache. 

c) \textit{High-staleness Synchronous Model Aggregation}: At the end of the training round, the high-staleness {customized} models stored in the cache are aggregated into a single global model, which is denoted as
\begin{align}\label{syn_aggre_single}
{\bf{W'}} = \sum\limits_{k \in \mathcal{L}} {{\varsigma ^k}{{\bf{W}}^k}}, 
\end{align}
where {${\varsigma ^k} \!=\! \frac{{{{\left| {\cal D}^k \right|}}}}{{{\left| {\cal D} \right|} }}$}, and $\mathcal{L}$ is the set of high-staleness models stored in the cache. After that, the model ${\bf{W'}}$ is aggregated into the latest model with smaller weights, which is represented as 
\begin{align}\label{syn_aggre}
{{\bf{W}}^*} = \beta {{\bf{W}}^*} + \left( {1 - \beta } \right){{\bf{W'}}}
\end{align}
where $\beta$ is the mixing hyperparameter of synchronous aggregation. {In the following, we use an example to illustrate this whole process of PMAS in Fig.~\ref{fig:asyn_agg_illu}.}


{Four inter-group sets are considered in this training round, and the initial latest model version is $L_0$.} 
When satellites in the first inter-group set establish a connection with a GS, {all basic sub-structure models} from the satellites are assembled into a customized global model $A_1$ via the sub-structure customization in Sec.~\ref{sec:sub_cus}. 
The customized model $A_1$ is classified as a high-staleness model,
thereby storing it in the cache rather than aggregating it into the latest model. When the satellites in the second and third {inter-group sets} establish connections with the GS, their customized models $A_2$ and $A_3$ are categorized as low-staleness models. Therefore, both customized models are asynchronously aggregated into the latest model. Afterward, as with $A_1$, the customized model $A_4$ from the fourth {inter-group set} is stored in the cache due to its high model staleness. In the end, all the customized models stored in the cache (i.e., $A_1$, $A_4$) are synchronously aggregated into the latest model with a {smaller weight.}

In short, unlike existing strategies that solely rely on time-dependent weighting functions for asynchronous aggregation, our strategy incorporates the similarity of models weights into the weighting function and asynchronously aggregates the low-staleness models in a timely manner to expedite model convergence, while storing the high-staleness models in the cache and aggregating them at the end of the training round to enhance generalization.

\section{Implementation and Experimental Setup} 
\label{sec:implementation}
{
\quad\quad\!\!\!\textit{a) Implementation}: We implement \name using Python 3.7 and PyTorch 1.9.1., and train it on a ThinkPad P17 Gen1 laptop equipped with an NVIDIA Quadro RTX 3000 GPU, Intel i9-10885H CPUs, and 4TB SSD. We employ the widely adopted and well-recognized VGG-16 network~\cite{simonyan2014very} in \name. VGG-16 is a classical deep convolutional neural network comprised of 13 convolution layers and 3 fully connected layers. It leverages the stacking of multiple convolution layers to effectively extract features from images, while the fully connected layers are responsible for classifying and predicting the extracted features. The learning rate for each satellite is uniformly set to 0.005, and the batch size is 128. The loss function and weighting function $ s\left( \tau  \right)$ are cross entropy loss and polynomial function~\cite{xie2019asynchronous}, respectively.


\textit{b) Experimental Setup}: In the experiments, we deploy $N$ satellites orbiting the Earth, distributed across $J$ different altitude orbits. $N$ and $J$ are set to 24 and 3 by default unless specified otherwise. The default orbital period ratio is  $p_1:p_2,...,:p_J=1:1.1:1.3$. The computing and memory resource budgets among satellites follow a discrete uniform distribution (i.e., $P(b_i=0.25, 0.5, 0.75, 1)=0.25$). The uplink rates are set based on real-world traces~(e.g., RTT) collected from Starlink.  

We adopt the space modulation recognition dataset GBSense~\cite{gbsense} and the remote sensing image dataset EuroSAT~\cite{helber2018introducing} to evaluate the training performance of \name. The GBSense consists of sampled signals with 13 modulation types and comprises 16000 training samples as well as 4000 test samples. EuroSAT contains 10 distinct categories of remote sensing images, such as industrial, highway, and forest. EuroSAT has 21600 training samples and 5400 test samples.
Furthermore, we conduct experiments under IID and non-IID data settings. In the IID setting, the training samples are randomly partitioned into 24 equal shards, with each shard assigned to one satellite. In the non-IID setting, we sort the data by labels, divide it into 240 shards, and distributed 10 shards to each of the 24 satellites.

}



\begin{figure}[t]
\setlength\abovecaptionskip{3pt}
\centering
\subfigure[{GBSense under IID setting.}]{
    \includegraphics[height=2.3cm,width=4.05cm]{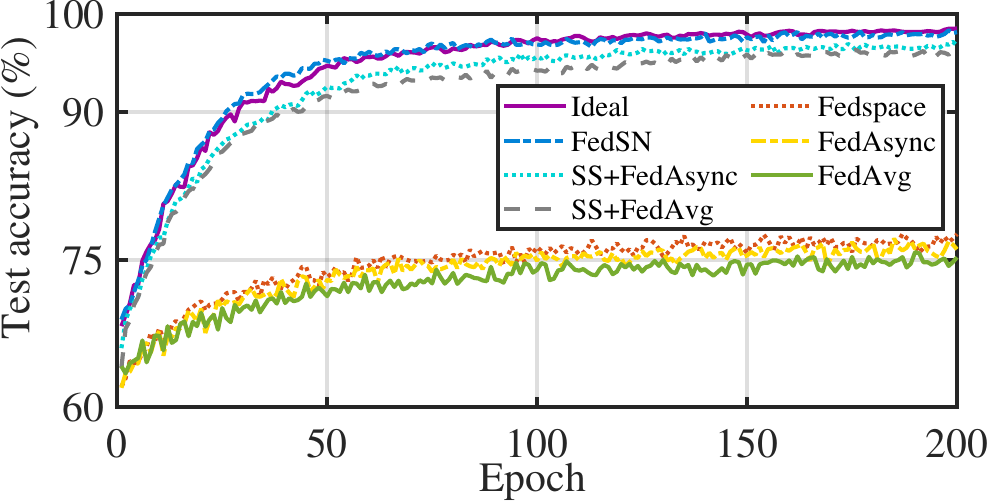}
    \label{sfig:Gbsense_iid}
}
\subfigure[{GBSense under non-IID setting.}]{
    \includegraphics[height=2.4cm,width=4cm]{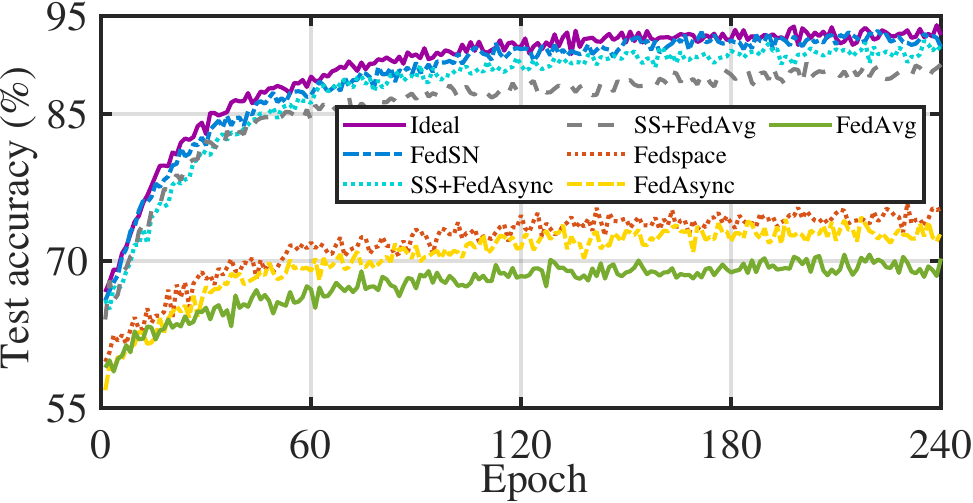}
    \label{sfig:Gbsense_non_iid}
}
\subfigure[{EuroSAT under IID setting.}]{
    \includegraphics[height=2.4cm,width=4cm]{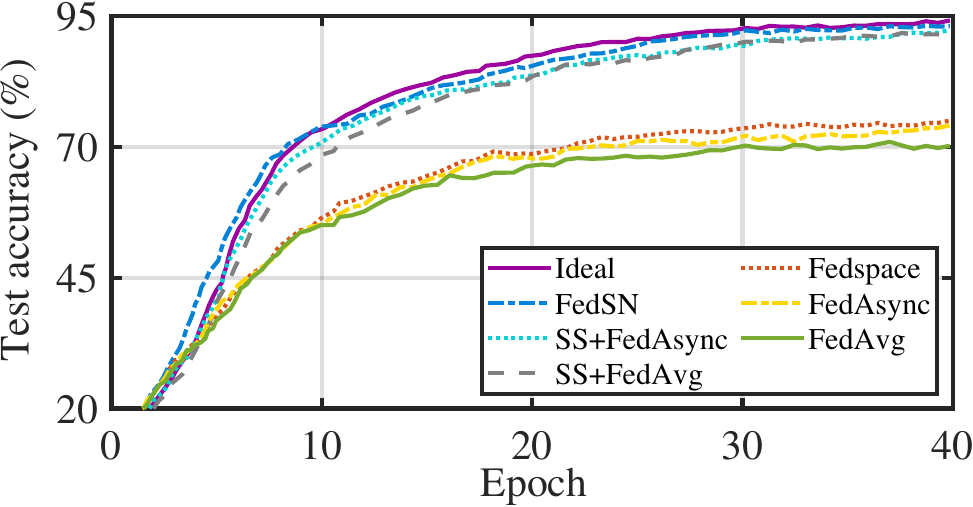}
    \label{sfig:EuroSAT_iid}
}
\subfigure[{EuroSAT under non-IID setting.}]{
    \includegraphics[height=2.4cm,width=4cm]{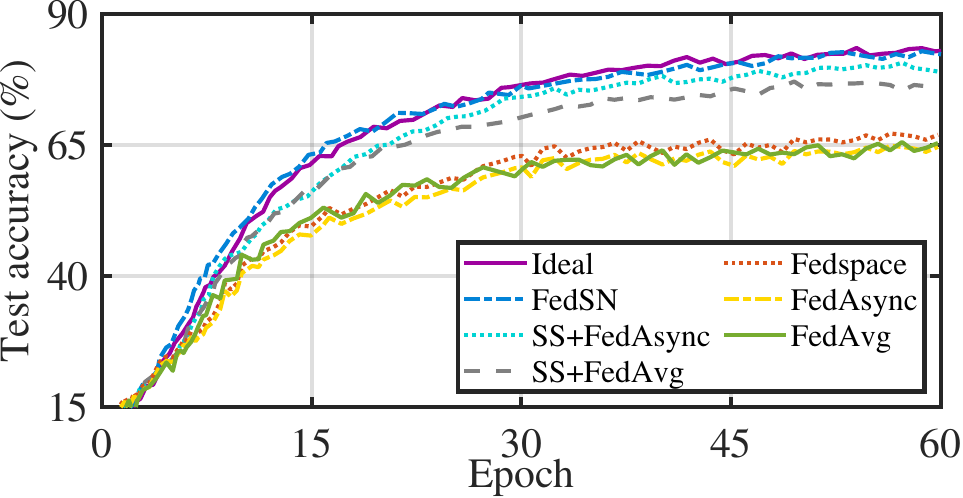}
    \label{sfig:EuroSAT_non_iid}
}
    \caption{ {The test accuracy for GBSense and EuroSAT dataset under IID and non-IID settings using VGG-16.}}
    \label{fig:overall_perfo}
    \vspace{-2ex}
\end{figure}

\section{Evaluation} \label{sec:evaluation}

{In this section, we provide extensive evaluations to demonstrate advantages of \name.}
We first assess the overall performance of \name, and 
conduct ablation experiments to illustrate effectiveness of each meticulously designed component in \name.
\vspace{-0.2cm}
\subsection{System Evaluation}

In this section, we conduct the overall performance evaluation of \name framework in terms of test accuracy, computing overhead, and communication overhead. Additionally, we also assess the performance for \name framework when customizing models with different widths and the impact of the number of satellites $N$ on training performance. To investigate the advantages of the \name framework, we compare it with five other benchmarks:
\begin{itemize}
    \item {\bf{Ideal:}} {The ideal case is resource- and staleness-unconstrained FedAvg~\cite{mcmahan2017communication}. In this ideal case, it is assumed that all participating satellites have sufficient resources (i.e., computing, storage, and communication) to perform timely model updates and transmissions for synchronous local model aggregation.}
    \item {\bf{SS+FedAsync:}} The SS+FedAsync benchmark utilizes the sub-structure scheme, as described in Sec.~\ref{sec:sub_scheme}, for intra-group model training. For inter-group model aggregation, the FedAsync~\cite{xie2019asynchronous} is employed with a polynomial weighting function.
    \item {\bf{SS+FedAvg:}} The SS+FedAvg benchmark deploys the sub-structure scheme for intra-group model training, and employs the FedAvg~\cite{mcmahan2017communication} for inter-group model aggregation.
    \item {{\bf{Fedspace:}}  A novel FL framework for satellite networks that schedules model aggregation based on the deterministic and time-varying connectivity according to satellite orbits~\cite{so2022fedspace}.}
    \item {\bf{FedAsync:}} The trained models from all satellites consistently are aggregated into the latest model in a asynchronous manner~\cite{xie2019asynchronous}.
    \item {\bf{FedAvg:}} The trained models from all satellites are stored in the GS and synchronously aggregated as an latest updated model at the end of each training round~\cite{mcmahan2017communication}. 
\end{itemize}

\begin{figure}[t]
\setlength\abovecaptionskip{3pt}
\centering
\subfigure[{GBSense dataset.}]{
    \includegraphics[height=2.25cm,width=4cm]{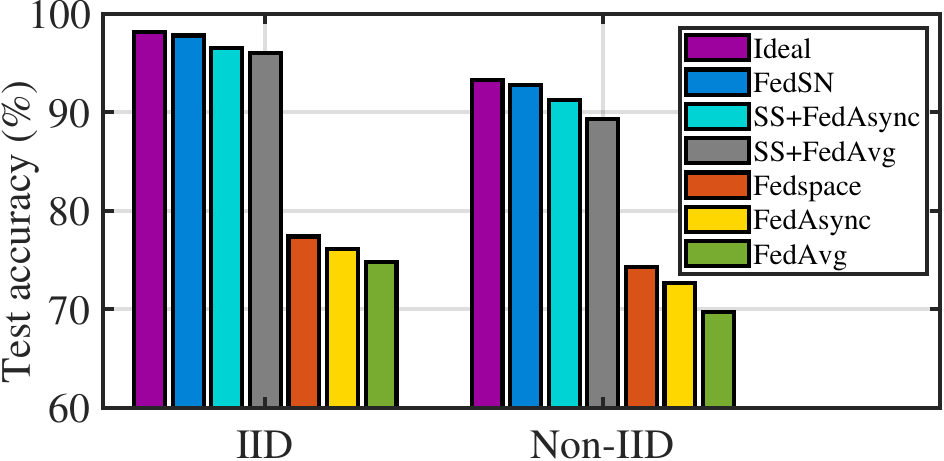}
    \label{sfig:gbsense_accuracy}
}
\subfigure[{EuroSAT dataset.}]{
    \includegraphics[height=2.25cm,width=4cm]{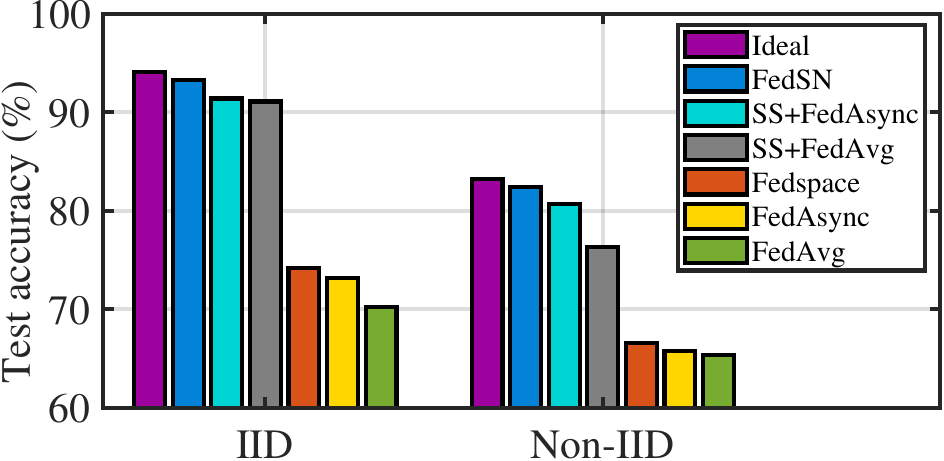}
    \label{sfig:euro_accuracy}
}

    \caption{{The converged test accuracy for GBSense and EuroSAT datasets under both IID and non-IID settings using VGG-16.}}
    \label{fig:gbsense_euro_accuracy}
    \vspace{-2ex}
\end{figure}

\subsubsection{The Overall Performance of \name}

Fig.~\ref{fig:overall_perfo} shows the test accuracy of \name and five other benchmarks on the GBSense and EuroSAT {datasets, respectively}. It is seen that \name retains test accuracy comparable to the ideal case as the model converges. {Apparently, \name outperforms SS+FedAsync, SS+FedAsync, Fedspace, FedAsync and FedAvg. It is worth noting that the test accuracy of FedSN, SS+FedAsync, and SS+FedAvg is notably superior to that of the Fedspace, FedAsync, and FedAvg, due to {deployment} of the sub-structure scheme.} This is because that sub-structure scheme incorporates three components to mitigate the adverse impacts of resource constraints, uneven training, and intra-group model staleness on the training process, respectively. Comparing \name with SS+FedAsync, reveals that the proposed pseudo-asynchronous model aggregation strategy can effectively compensate for inter-group staleness {caused by} model version discrepancies, thus improving the test accuracy in model training. {Furthermore, the comparisons of Fig.~\ref{sfig:Gbsense_iid} and Fig.~\ref{sfig:Gbsense_non_iid}, Fig.~\ref{sfig:EuroSAT_iid} and Fig.~\ref{sfig:EuroSAT_non_iid} show that the convergence speed of \name and five other benchmarks is slower under non-IID setting than under IID setting.}

\begin{figure}[t]
\setlength\abovecaptionskip{3pt}
\centering
\subfigure[{Computing overhead on GBSense.}]{
\includegraphics[height=2.45cm,width=4cm]{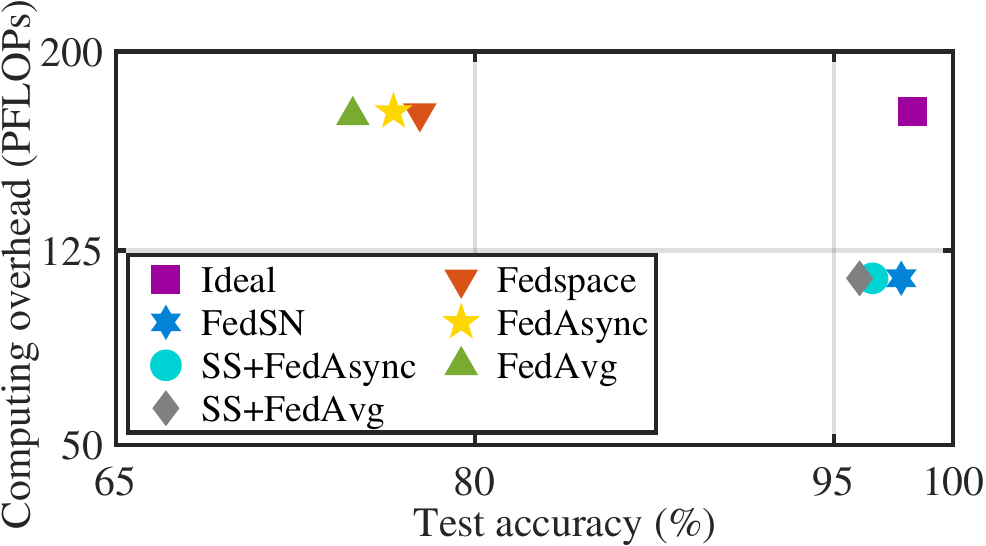}
    \label{sfig:gbsense_compute_accuracy}
}
\subfigure[{Communication overhead on GBSense.}]{
    \includegraphics[height=2.4cm,width=4cm]{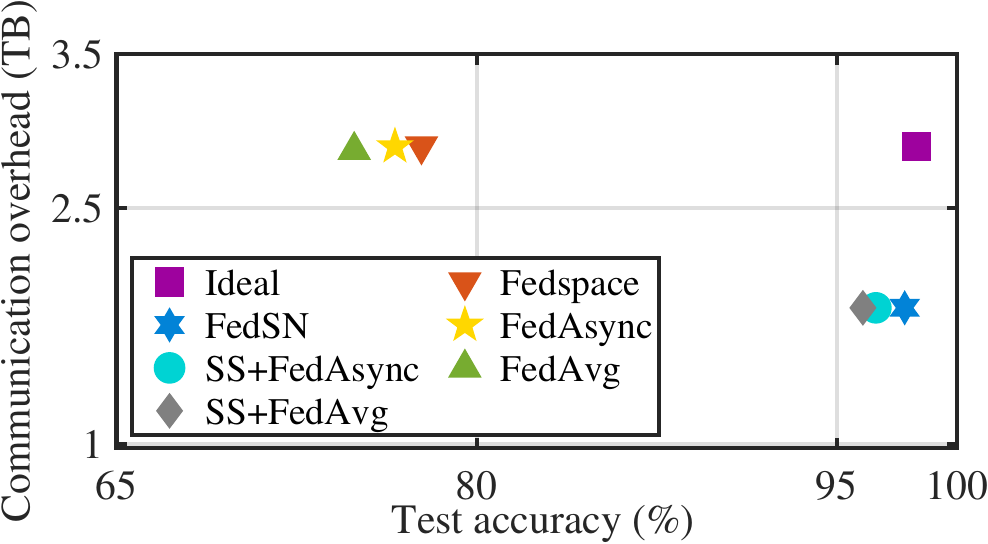}
    \label{sfig:gbsense_communication_accuracy}
}
\subfigure[{Computing overhead on EuroSAT.}]{
\includegraphics[height=2.45cm,width=4cm]{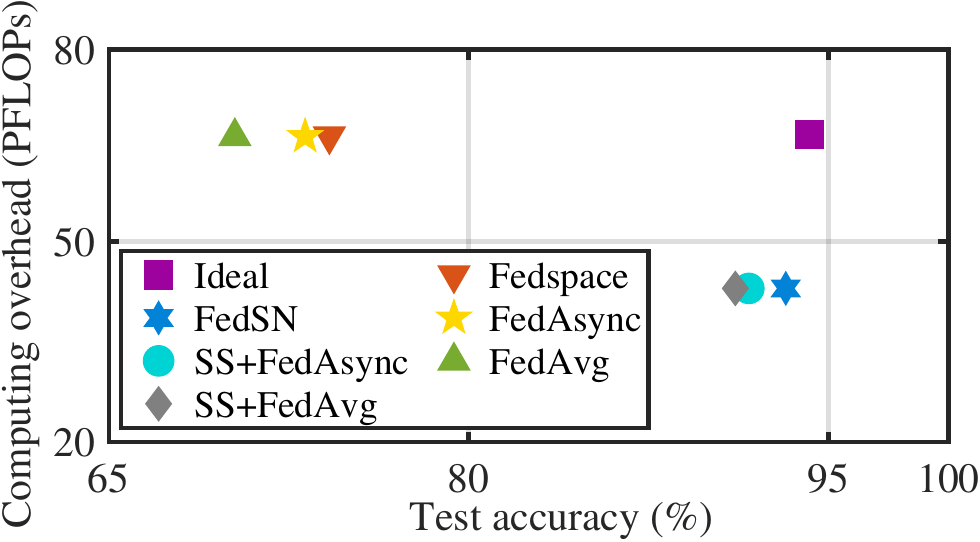}
    \label{sfig:euro_compute_accuracy}
}
\subfigure[{Communication overhead on EuroSAT.}]{
    \includegraphics[height=2.4cm,width=4cm]{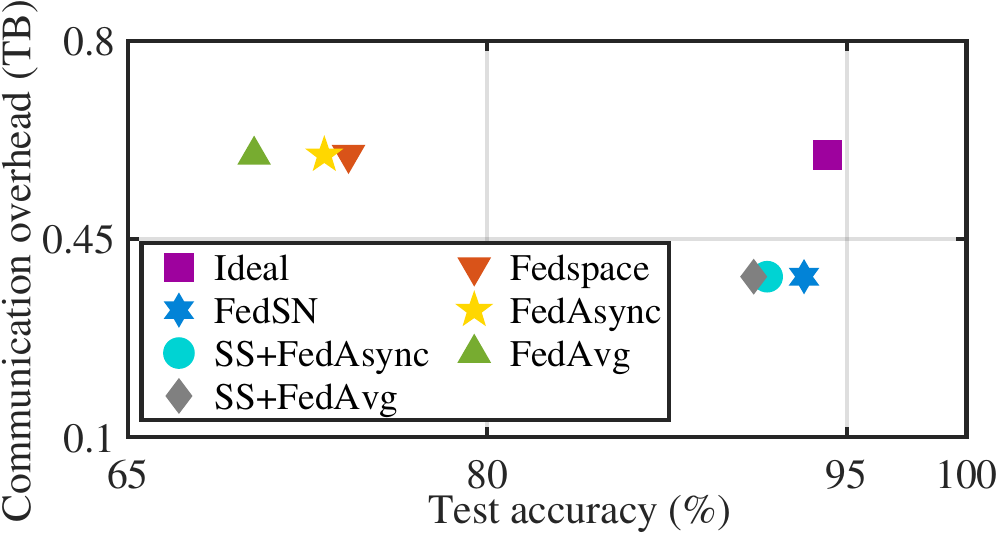}
    \label{sfig:euro_communication_accuracy}
}
    \caption{{The total computing and communication overhead for GBSense and EuroSAT datasets under IID settings using VGG-16.}}
    \label{fig:compute_communication_accuracy}
    \vspace{-2ex}
\end{figure}

Fig.~\ref{fig:gbsense_euro_accuracy} presents the converged test accuracy for GBSense and EuroSAT datasets, where the accuracy is taken at the 250-th epoch (GBSense) and 70-epoch (EuroSAT) when these learning frameworks all converge. Fig.~\ref{sfig:gbsense_accuracy} and Fig.~\ref{sfig:euro_accuracy} demonstrate that \name achieves test accuracy of 98.19$\%$ and 92.88$\%$ (94.11$\%$ and 83.26$\%$) on the GBSense 
(EuroSAT) dataset under IID and non-IID settings, with a negligible  decrease of only 0.54$\%$ and 0.78$\%$ (0.79$\%$ and 0.8$\%$) compared to the Ideal case. This indicates that although \name cannot provide global model training for each satellite due to resource constraints, the well-designed sub-structure scheme mitigates this issue and integrates more useful information dispersed across satellites into the model training, resulting in training performance comparable to the ideal case. Furthermore, by comparing \name with SS+FedAvg and SS+FedAsync, we observe the superiority of the proposed pseudo-synchronous model aggregation strategy over conventional strategies. In the case of Fedavg and FedAsync, as they are not specifically tailored to address resource heterogeneity, some satellites are consistently excluded from model training, leading to poorer generalization. In addition, Fig.~\ref{sfig:gbsense_accuracy} and Fig.~\ref{sfig:euro_accuracy} show that the converged test accuracy of \name and five other benchmarks is higher under non-IID setting than under IID setting.

\subsubsection{The Computing and Communciation Overhead}

Fig.~\ref{fig:compute_communication_accuracy} illustrates the total computing overhead, and communication overhead for GBSense and EuroSAT datasets. It can be seen that \name framework reduces computing and communication overhead by approximately 38$\%$ while retaining similar test accuracy as the ideal case. The reason is twofold: one is that the sub-structure scheme significantly reduces the number of network parameters by channel-wise partitioning, thus reducing the overall computing and communication overhead. The other is that the sub-structure scheme effectively mitigates information loss caused by resource constraints, uneven training, and intra-group model staleness. And the pseudo-asynchronous model aggregation strategy eliminates the adverse effects of inter-group model staleness by designing weighting functions as well as a buffer-based model aggregation scheme. Overall, FedSN is a more communication- and compute-efficient framework with marginal accuracy loss.

\begin{figure}[t]
\setlength\abovecaptionskip{3pt}
\centering
\subfigure[GBSense under IID setting.]{
    \includegraphics[height=2.4cm,width=4cm]{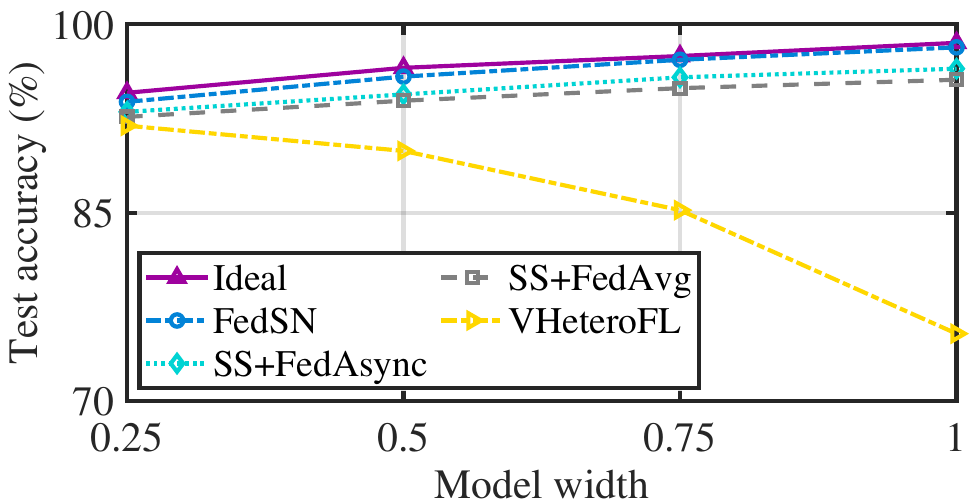}
    \label{sfig:width_Gbsense_iid}
}
\subfigure[GBSense under non-IID setting.]{
    \includegraphics[height=2.4cm,width=4cm]{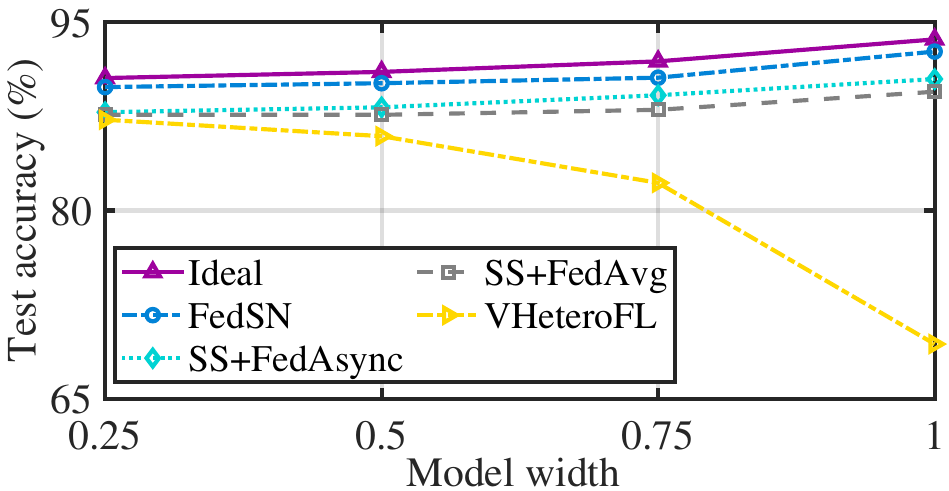}
    \label{sfig:width_Gbsense_non_iid}
}
\subfigure[EuroSAT under IID setting.]{
    \includegraphics[height=2.4cm,width=4cm]{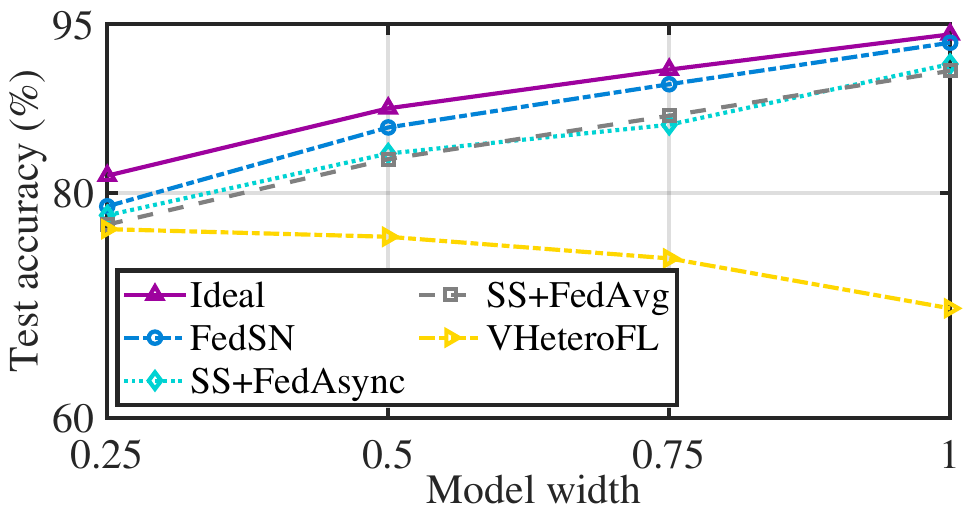}
    \label{sfig:width_EuroSAT_iid}
}
\subfigure[EuroSAT under non-IID setting.]{
    \includegraphics[height=2.4cm,width=4cm]{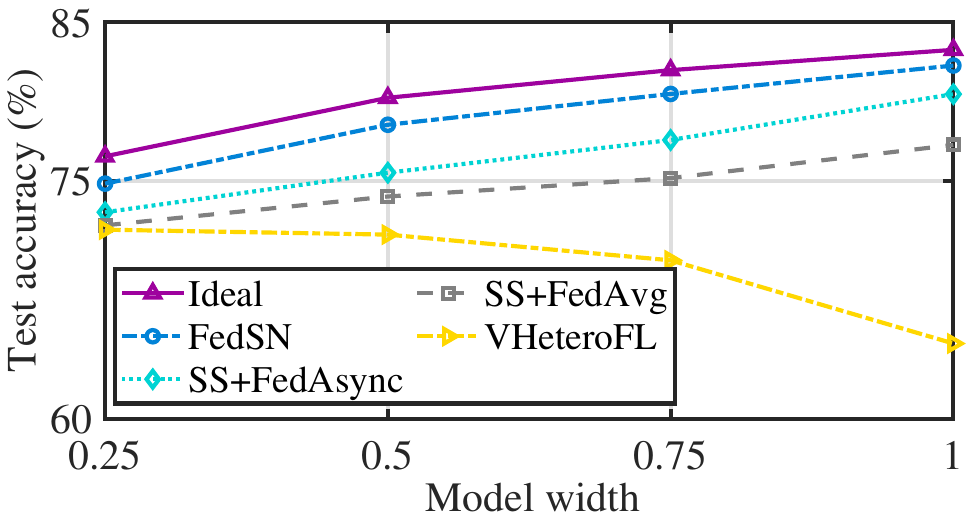}
    \label{sfig:width_EuroSAT_non_iid}
}
    \caption{ The converged test accuracy versus the customized model width on GBSense and EuroSAT datasets under IID and non-IID settings using VGG-16.}
    \label{fig:width_overall_perfo}
    \vspace{-2ex}
\end{figure}

\subsubsection{The Performance of \name for Different Customized Model Widths}

Fig.~\ref{fig:width_overall_perfo} shows the converged test accuracy versus the customized model width on GBSense and EuroSAT datasets. For VHeteroFL, we only present the performance evaluated on trained widths. Obviously, the proposed FedSN framework always achieves similar test accuracy as the ideal case for customized models with different widths,  thereby demonstrating the excellent scalability of the FedSN framework. As for VHeteroFL, satellites' constrained resources hinder wider models from fully leveraging the dispersed data, leading to inferior  training performance. In contrast, FedSN overcomes the under-training issue caused by constrained resources via partitioning-assembling principle. To be specific, the global model is first partitioned into basic sub-structure models according to the minimum budget constraint, and then re-assembles them into the customized global model. Consequently, FedSN outperforms VHeteroFL in terms of test accuracy at each of model widths.

\subsubsection{The Impact of Number of Participating Satellites}

\begin{figure}[t]
\setlength\abovecaptionskip{3pt}
\centering
\subfigure[GBSense under IID setting.]{
    \includegraphics[height=2.4cm,width=4cm]{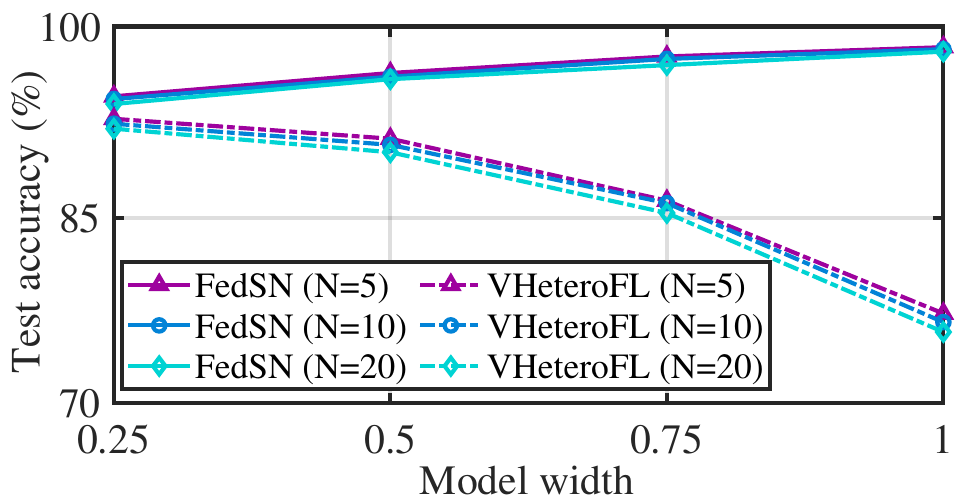}
    \label{sfig:number_Gbsense_iid}
}
\subfigure[GBSense under non-IID setting.]{
    \includegraphics[height=2.4cm,width=4cm]{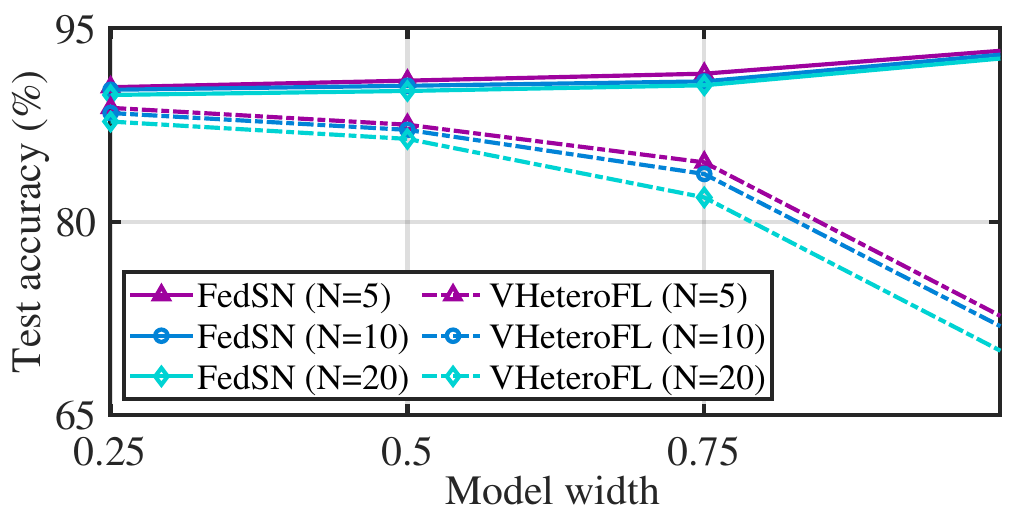}
    \label{sfig:number_Gbsense_non_iid}
}
\subfigure[EuroSAT under IID setting.]{
    \includegraphics[height=2.4cm,width=4cm]{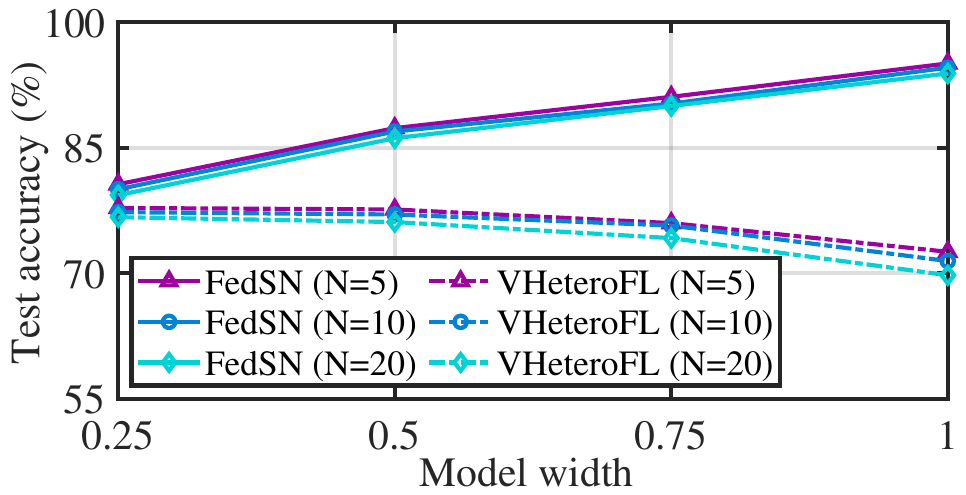}
    \label{sfig:number_EuroSAT_iid}
}
\subfigure[EuroSAT under non-IID setting.]{
    \includegraphics[height=2.4cm,width=4cm]{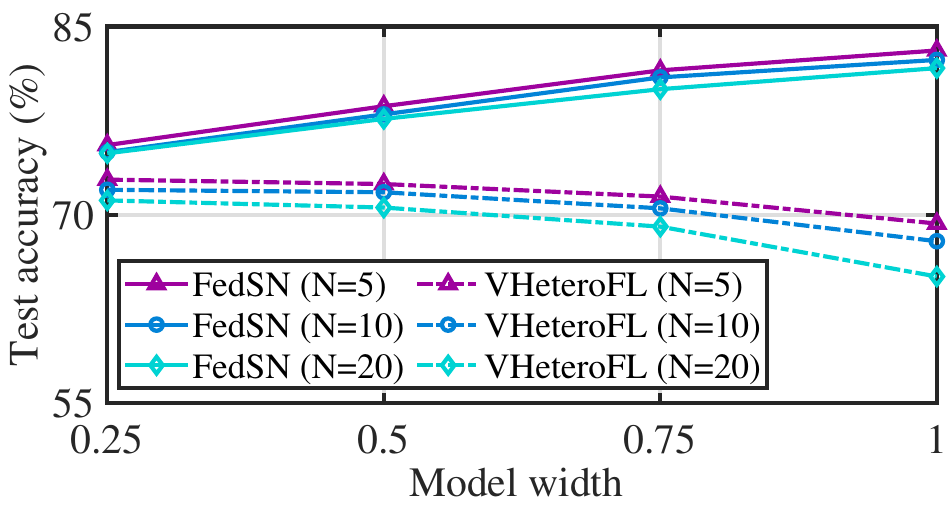}
    \label{sfig:number_EuroSAT_non_iid}
}
    \caption{ The impact of the number of participating satellites $N$ on converged test accuracy on GBSense and EuroSAT datasets under both IID and non-IID settings using VGG-16.}
    \label{fig:number_overall_perfo}
    \vspace{-2ex}
\end{figure}

Fig.~\ref{fig:number_overall_perfo} presents the impact of the number of participating satellites on converged test accuracy for GBSense and EuroSAT datasets. It is observed that the curves for all satellite numbers follow a similar trend in each plot. The experimental results for satellite ranging from 5 to 20 show a negligible effect on the performance of FedSN, whereas VHeteroFL exhibits a decrease in test accuracy as the number of participating satellites increases, thus highlighting the superior robustness of FedSN over VHeteroFL. With an increase in the number of satellites, the distribution of resource constraints becomes more heterogeneous. For VHeteroFL, the fixed sub-structure model customization deters it from handling the high heterogeneity of constrained resources effectively. In contrast, FedSN's sub-structure model customization enables it to overcome the heterogeneous resources by flexible sub-structure partitioning and assembling, resulting in enhanced robustness. Furthermore, we observe that training performance of FedSN consistently outperforms VHeteroFL across various numbers of satellites.

\subsection{Micro-benchmark}
This subsection conducts ablation experiments to demonstrate the effectiveness of each elaborately designed component (i.e., sub-structure customization, sub-structure distribution, sub-structure aggregation, and pseudo-synchronous model aggregation.) in \name.

\subsubsection{Sub-structure Customization}

Fig.~\ref{sfig:sub_cus_ablation} shows the impact of the sub-structure customization method on training performance for the GBsense dataset. In the case of FedAvg, the resource constraints on satellites deter the effective integration of dispersed information into the trained model, resulting in the worst training performance. However, by deploying the proposed sub-structure customization method on FedAvg, satellites can leverage more local residing data to train their basic sub-structure combinations, resulting in a significant improvement in test accuracy from 75.3$\%$ to 94.4$\%$. Furthermore, the relatively small performance gap observed between Ideal and SCM+FedAvg further corroborates the effectiveness of the proposed sub-structure customization method.

\subsubsection{Sub-structure Distribution}

Fig.~\ref{sfig:sub_dis_ablation} illustrates the impact of different sub-structure distribution methods on training performance for the GBsense dataset. SCM+RD and SCM+SD exhibit lower test accuracy due to the random and fixed selection of sub-structure combinations, which leads to uneven training of different parts of the global model. The comparison between SCM+SDM, SCM+RD, and SCM+SD highlights the superiority  of our proposed adaptive scrolling sub-structure distribution method, which better ensures more even model training, thus leading to higher training accuracy.

\subsubsection{Sub-structure Aggregation}

\begin{figure}[t]
\setlength\abovecaptionskip{3pt}
\centering
\subfigure[Sub-structure customization.]{
    \includegraphics[height=2.4cm,width=4cm]{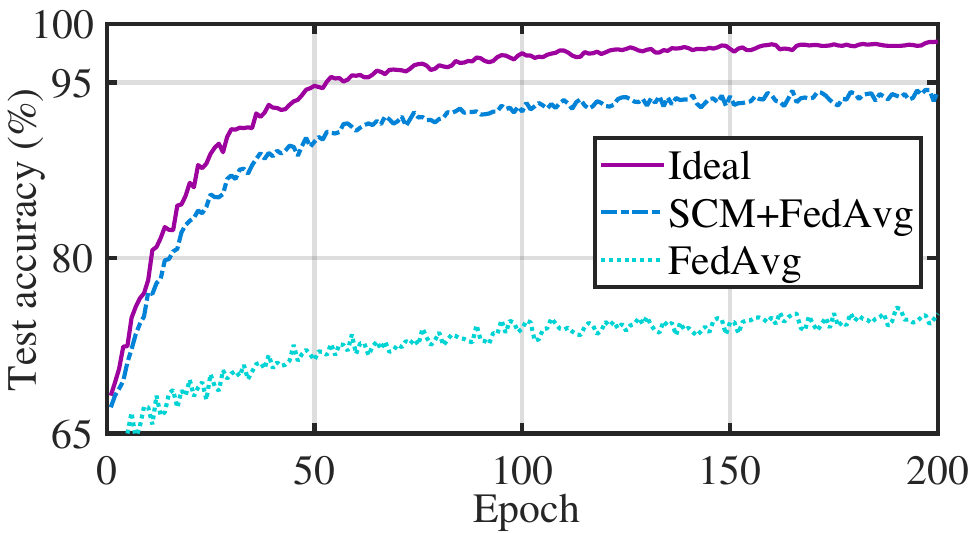}
    \label{sfig:sub_cus_ablation}
}
\subfigure[Sub-structure distribution.]{
    \includegraphics[height=2.4cm,width=4cm]{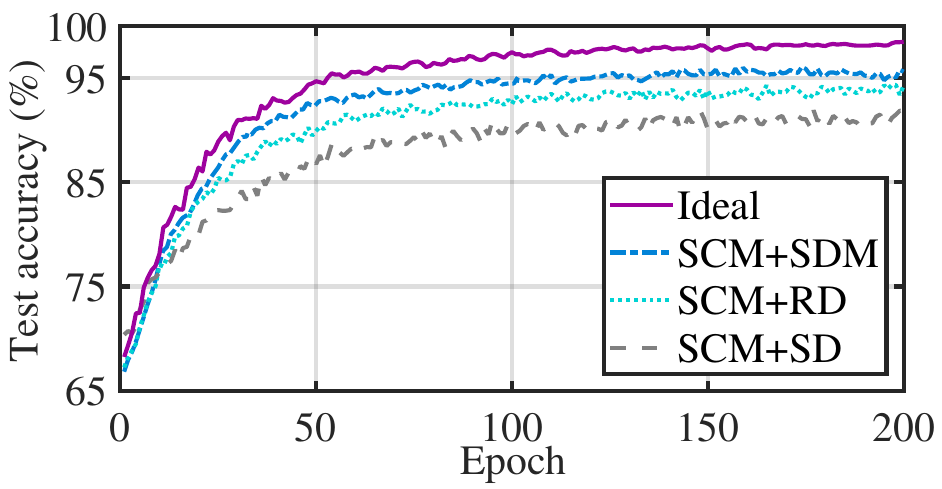}
    \label{sfig:sub_dis_ablation}
}
\subfigure[Sub-structure aggregation.]{
    \includegraphics[height=2.4cm,width=4cm]{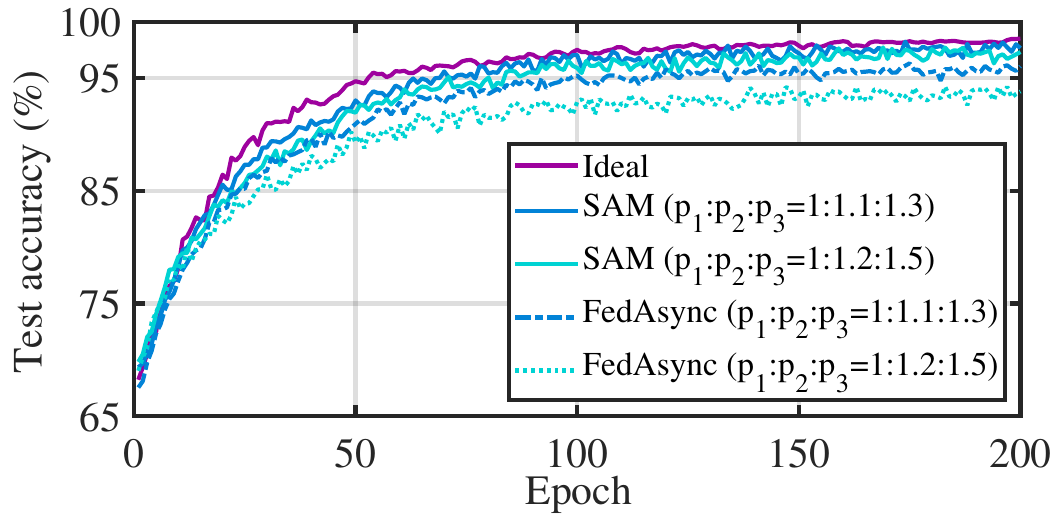}
    \label{sfig:sub_agg_ablation}
}
\subfigure[Pseudo-synchronous model aggregation.]{
    \includegraphics[height=2.4cm,width=4cm]{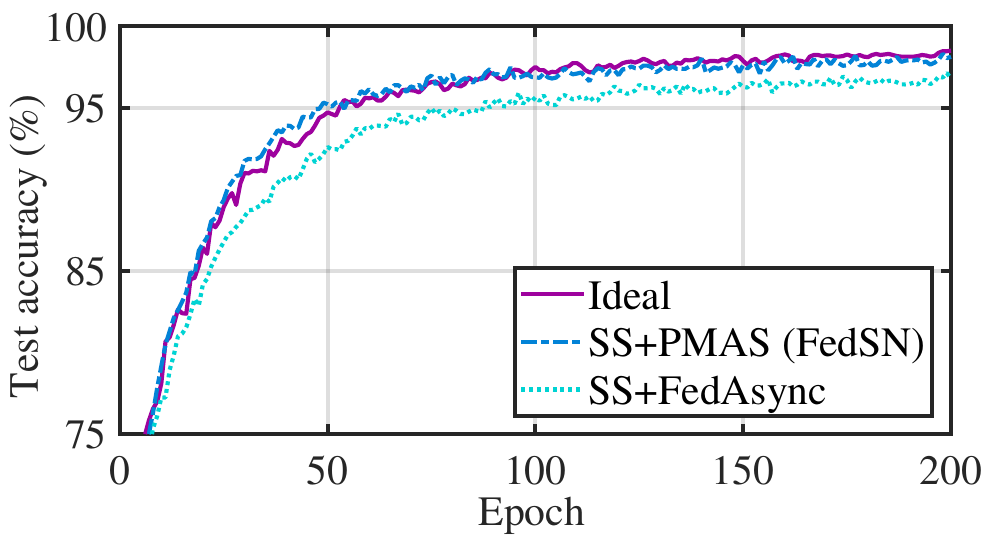}
    \label{sfig:sub_interagg_ablation}
}
    \caption{Ablation experiments for sub-structure customization\,(a), sub-structure distribution\,(b), sub-structure aggregation\,(c), and pseudo-synchronous model aggregation\,(d) on the GBsense dataset under IID setting.}
    \label{fig:ablation_experiment}
    \vspace{-2ex}
\end{figure}

Fig.~\ref{sfig:sub_agg_ablation} presents the impact of the sub-structure aggregation method on training performance for the GBsense dataset. The comparison between SAM and FedAvg demonstrates the effectiveness of our proposed sub-structure aggregation method in mitigating the intra-group model staleness arising from orbit period differences, thus improving the generalization of model training. This is achieved by modifying the aggregation weights to assign larger weights to satellites with longer orbit periods and smaller weights to those with shorter orbit periods, thereby ensuring balanced participation of satellites

\subsubsection{Pseudo-synchronous Model Aggregation}

Fig.~\ref{sfig:sub_interagg_ablation} demonstrates the impact of various inter-group aggregation strategies on training performance for the GBsense dataset. By comparing SS+PMAS and SS+FedAsync, it is evident that the proposed customized pseudo-asynchronous model aggregation strategy significantly expedites model convergence and improves test accuracy. There are two reasons for this: one is that the proposed strategy asynchronously aggregates low-staleness customized models promptly to speed up model convergence. The other is that our strategy aggregates high-staleness customized models with smaller weights at the end of the training round to enhance the generalization.

\section{Related Work} \label{sec:related_work}

FL has been widely regarded as the prominent distributed learning paradigm, primarily due to its advantages in communication efficiency and data privacy. Unfortunately, research on FL in satellite networks is still in its infancy. The prominent FL frameworks tailored for satellite networks focus their efforts on tackling two critical challenges: One is model staleness caused by intermittent connectivity~\cite{wu2023fedgsm,elmahallawy2022asyncfleo,razmi2022scheduling,so2022fedspace}. To alleviate this issue, Wu~\textit{et al.}~\cite{wu2023fedgsm} propose an asynchronous model aggregation strategy that utilizes the discrepancy between local updates to offset the adverse effects of model staleness, thereby improving the accuracy and robustness of model training. To achieve fast convergence speed and high test accuracy, Elmahallawy~\textit{et al.}~\cite{elmahallawy2022asyncfleo} develop an efficient asynchronous FL framework featuring a model aggregation strategy that incorporates satellite grouping and staleness discounting. Razmi~\textit{et al.}~\cite{razmi2022scheduling} present an optimal time scheduling scheme leveraging the predictability of visiting time between GS and satellites to mitigate model staleness. So~\textit{et al.}~\cite{so2022fedspace} formulate an optimization problem to capture the fundamental trade-off between idle connectivity and local model staleness, and proposes an innovative FL framework based on the deterministic and time-varying connectivity in satellite orbits. However, existing asynchronous FL strategies solely utilize time-dependent weighting functions to balance the previous and current models, which ignores discrepancies between model weights and thus does not perfectly characterize the extent of model staleness. 

The other is heterogeneous connectivity among satellites~\cite{elmahallawy2022fedhap}. To alleviate the detrimental effects of this issue on FL performance,  Elmahallawy~\textit{et al.}~\cite{elmahallawy2022fedhap} devise a novel FL framework based on inter-satellite collaboration, which addresses the challenges of highly sporadic and heterogeneous satellite-GS connectivity in LEO constellations through the design of a hierarchical communication architecture, model  dissemination scheme, and model aggregation scheme. Nevertheless, this approach requires the deployment of additional equipment, leading to increased deployment costs and inter-satellite or HAP-ground communication overhead.

In practice, different satellites have various on-device resources, and their allocated resources for FL training may be drastically changed during run-time, which causes under-training of satellites and thus leads to model bias. Furthermore, the difference in orbital periods  among satellites at different altitudes results in imbalanced participation in model training, thereby lowering the effectiveness of the training process. However, to the best of our knowledge, there is no existing work that has developed FL frameworks considering resource heterogeneity and orbital period differences.

\section{Conclusion} \label{sec: conclusion}

In this paper, we {propose} a novel FL framework over LEO satellite networks, named \name, to enhance the effectiveness of model training. \name consists of two main components: sub-structure scheme and pseudo-synchronous model aggregation. The sub-structure scheme devises the sub-structure customization, distribution, and aggregation methods to address the resource limitations, training imbalance, and intra-group model staleness, respectively. The pseudo-synchronous model aggregation strategy incorporates discrepancies between model weights into a weighting function and develops a buffer-based aggregation method to mitigate inter-group model staleness.  Extensive experimental results demonstrate that \name framework outperforms the state-of-the-art benchmarks. 
This work has demonstrated the potential of deploying
\name over LEO satellite networks. In the future, we would like to leverage model quantization and pruning to further address bottlenecks of uplink communications.



%





\ifCLASSOPTIONcaptionsoff
  \newpage
\fi



%



\bibliographystyle{IEEEtran}
\bibliography{reference}

\begin{thebibliography}{10}
\providecommand{\url}[1]{#1}
\csname url@samestyle\endcsname
\providecommand{\newblock}{\relax}
\providecommand{\bibinfo}[2]{#2}
\providecommand{\BIBentrySTDinterwordspacing}{\spaceskip=0pt\relax}
\providecommand{\BIBentryALTinterwordstretchfactor}{4}
\providecommand{\BIBentryALTinterwordspacing}{\spaceskip=\fontdimen2\font plus
\BIBentryALTinterwordstretchfactor\fontdimen3\font minus \fontdimen4\font\relax}
\providecommand{\BIBforeignlanguage}[2]{{%
\expandafter\ifx\csname l@#1\endcsname\relax
\typeout{** WARNING: IEEEtran.bst: No hyphenation pattern has been}%
\typeout{** loaded for the language `#1'. Using the pattern for}%
\typeout{** the default language instead.}%
\else
\language=\csname l@#1\endcsname
\fi
#2}}
\providecommand{\BIBdecl}{\relax}
\BIBdecl

\bibitem{mark2018tech}
H.~Mark, ``{Tech Giants Race to Build Orbital Internet},'' \emph{IEEE Spectr.}, vol.~55, no.~6, pp. 10--11, 2018.

\bibitem{ahmmed2022digital}
T.~Ahmmed, A.~Alidadi, Z.~Zhang, A.~U. Chaudhry, and H.~Yanikomeroglu, ``{The Digital Divide in Canada and the Role of LEO Satellites in Bridging the Gap},'' \emph{{IEEE} Commun. Mag.}, vol.~60, no.~6, pp. 24--30, 2022.

\bibitem{chen2022robust}
Q.~Chen, W.~Meng, S.~Han, C.~Li, and H.-H. Chen, ``{Robust Task Scheduling for Delay-aware IoT Applications in Civil Aircraft-augmented SAGIN},'' \emph{{IEEE} Trans. Commun.}, vol.~70, no.~8, pp. 5368--5385, 2022.

\bibitem{wu2024accelerating}
J.~Wu, S.~Su, X.~Wang, J.~Zhang, and Y.~Gao, ``Accelerating handover in mobile satellite network,'' in \emph{Proc. INFOCOM}, 2024.

\bibitem{yuan2023graph}
H.~Yuan, Z.~Chen, Z.~Lin, J.~Peng, Z.~Fang, Y.~Zhong, Z.~Song, X.~Wang, and Y.~Gao, ``{Graph Learning for Multi-satellite based Spectrum Sensing},'' in \emph{Proc. ICCT}, 2023.

\bibitem{aragon2018cubesats}
B.~Aragon, R.~Houborg, K.~Tu, J.~B. Fisher, and M.~McCabe, ``{CubeSats Enable High Spatiotemporal Retrievals of Crop-water Use for Precision Agriculture},'' \emph{Remote Sens.}, vol.~10, no.~12, p. 1867, 2018.

\bibitem{franch2020spatial}
I.~Franch-Pardo, B.~M. Napoletano, F.~Rosete-Verges, and L.~Billa, ``{Spatial Analysis and GIS in the Study of COVID-19. A Review},'' \emph{Sci. Total Environ.}, vol. 739, p. 140033, 2020.

\bibitem{shukla2021enhancing}
S.~Shukla, D.~Macharia, G.~J. Husak, M.~Landsfeld, C.~L. Nakalembe, S.~L. Blakeley, E.~C. Adams, and J.~Way-Henthorne, ``{Enhancing Access and Usage of Earth Observations in Environmental Decision-making in Eastern and Southern Africa Through Capacity Building},'' \emph{Front. Sustainable Food Syst.}, vol.~5, p. 504063, 2021.

\bibitem{letaief2021edge}
K.~B. Letaief, Y.~Shi, J.~Lu, and J.~Lu, ``{Edge Artificial Intelligence for 6G: Vision, Enabling Technologies, and Applications},'' \emph{{IEEE} J. Sel. Areas Commun.}, vol.~40, no.~1, pp. 5--36, 2021.

\bibitem{chen2021rf}
Z.~Chen, C.~Cai, T.~Zheng, J.~Luo, J.~Xiong, and X.~Wang, ``{Rf-based Human Activity Recognition Using Signal Adapted Convolutional Neural Network},'' \emph{{IEEE} Trans. Mobile Comput.}, vol.~22, no.~1, pp. 487--499, 2021.

\bibitem{lyu2023optimal}
S.~Lyu, Z.~Lin, G.~Qu, X.~Chen, X.~Huang, and P.~Li, ``{Optimal Resource Allocation for U-shaped Parallel Split Learning},'' in \emph{Proc. GlobeCom Wkshps}, 2023.

\bibitem{fang2024automated}
Z.~Fang, Z.~Lin, Z.~Chen, X.~Chen, Y.~Gao, and Y.~Fang, ``{Automated Federated Pipeline For Parameter-efficient Fine-tuning of Large Language Models},'' \emph{arXiv preprint arXiv:2404.06448}, 2024.

\bibitem{lin2024splitlora}
Z.~Lin, X.~Hu, Y.~Zhang, Z.~Chen, Z.~Fang, X.~Chen, A.~Li, P.~Vepakomma, and Y.~Gao, ``{SplitLoRA: A Split Parameter-efficient Fine-tuning Framework For Large Language Models},'' \emph{arXiv preprint arXiv:2407.00952}, 2024.

\bibitem{so2022fedspace}
J.~So, K.~Hsieh, B.~Arzani, S.~Noghabi, S.~Avestimehr, and R.~Chandra, ``{Fedspace: An Efficient Federated Learning Framework at Satellites and Ground Stations},'' \emph{arXiv preprint arXiv:2202.01267}, 2022.

\bibitem{lin2023efficient}
Z.~Lin, G.~Zhu, Y.~Deng, X.~Chen, Y.~Gao, K.~Huang, and Y.~Fang, ``{Efficient Parallel Split Learning over Resource-constrained Wireless Edge Networks},'' \emph{{IEEE} Trans. Mobile Comput.}, 2024.

\bibitem{mcmahan2017communication}
B.~McMahan, E.~Moore, D.~Ramage, S.~Hampson, and B.~A. y~Arcas, ``{Communication-efficient Learning of Deep Networks from Decentralized Data},'' in \emph{Proc. AISTATS}, 2017.

\bibitem{chen2020joint}
M.~Chen, Z.~Yang, W.~Saad, C.~Yin, H.~V. Poor, and S.~Cui, ``{A Joint Learning and Communications Framework for Federated Learning over Wireless Networks},'' \emph{{IEEE} Trans. Wireless Commun.}, vol.~20, no.~1, pp. 269--283, 2020.

\bibitem{xu2020client}
J.~Xu and H.~Wang, ``{Client Selection and Bandwidth Allocation in Wireless Federated Learning Networks: A Long-term Perspective},'' \emph{{IEEE} Trans. Wireless Commun.}, vol.~20, no.~2, pp. 1188--1200, 2020.

\bibitem{zhang2024fedac}
Y.~Zhang, H.~Chen, Z.~Lin, Z.~Chen, and J.~Zhao, ``{FedAC: A Adaptive Clustered Federated Learning Framework for Heterogeneous Data},'' \emph{arXiv preprint arXiv:2403.16460}, 2024.

\bibitem{lin2024adaptsfl}
Z.~Lin, G.~Qu, W.~Wei, X.~Chen, and K.~K. Leung, ``{AdaptSFL: Adaptive Split Federated Learning in Resource-constrained Edge Networks},'' \emph{arXiv preprint arXiv:2403.13101}, 2024.

\bibitem{handley2019using}
M.~Handley, ``{Using Ground Relays for Low-latency Wide-area Routing in Megaconstellations},'' in \emph{Proc. ACM Workshop HotNets}, 2019.

\bibitem{razmi2022ground}
N.~Razmi, B.~Matthiesen, A.~Dekorsy, and P.~Popovski, ``{Ground-assisted Federated Learning in LEO Satellite Constellations},'' \emph{{IEEE} Wireless Commun. Lett.}, vol.~11, no.~4, pp. 717--721, 2022.

\bibitem{wu2023fedgsm}
L.~Wu and J.~Zhang, ``{FedGSM: Efficient Federated Learning for LEO Constellations with Gradient Staleness Mitigation},'' \emph{arXiv preprint arXiv:2304.08537}, 2023.

\bibitem{elmahallawy2022asyncfleo}
M.~Elmahallawy and T.~Luo, ``{AsyncFLEO: Asynchronous Federated Learning for LEO Satellite Constellations with High-Altitude Platforms},'' \emph{arXiv preprint arXiv:2212.11522}, 2022.

\bibitem{razmi2022scheduling}
{Razmi, Nasrin and Matthiesen, Bho and Dekorsy, Armin and Popovski, Petar}, ``{Scheduling for Ground-Assisted Federated Learning in LEO Satellite Constellations},'' in \emph{Proc. EUSIPCO}, 2022.

\bibitem{gbsense}
``{{Deep Sub-Nyquist Modulation Recognition Challenge}}". Available: http://www.gbsense.net/challenge/.

\bibitem{simonyan2014very}
K.~Simonyan and A.~Zisserman, ``{Very Deep Convolutional Networks for Large-scale Image Recognition},'' in \emph{Proc. ICLR}, 2015.

\bibitem{haque2019object}
M.~F. Haque, H.-Y. Lim, and D.-S. Kang, ``{Object Detection Based on VGG with ResNet Network},'' in \emph{Proc. ICEIC}, 2019.

\bibitem{tirumala1999iperf}
A.~Tirumala, ``{Iperf: The TCP/UDP Bandwidth Measurement Tool},'' \emph{http://dast. nlanr. net/Projects/Iperf/}, 1999.

\bibitem{Devaraj2019PlanetHS}
K.~Devaraj, M.~Ligon, E.~Blossom, J.~Breu, B.~Klofas, K.~Colton, and R.~W. Kingsbury, ``{Planet High Speed Radio: Crossing Gbps from a 3U CubeSat},'' in \emph{Proc. Small Satellite Conference}, 2019.

\bibitem{telesat}
{D. Aneesh}, ``{{The Right Way to Introduce LEO Services}}". Available: https://www.telesat.com/blog/the-right-way-to-introduce-leo-services/, 2023.

\bibitem{vasisht2021l2d2}
D.~Vasisht, J.~Shenoy, and R.~Chandra, ``{L2D2: Low Latency Distributed Downlink for LEO Satellites},'' in \emph{Proc. ACM SIGCOMM}, 2021.

\bibitem{hubert2002linux}
B.~Hubert \emph{et~al.}, ``{Linux Advanced Routing \& Traffic Control HOWTO},'' \emph{Netherlabs BV}, vol.~1, pp. 99--107, 2002.

\bibitem{duggen2007subduction}
S.~Duggen, P.~Croot, U.~Schacht, and L.~Hoffmann, ``{Subduction Zone Volcanic Ash Can Fertilize The Surface Ocean and Stimulate Phytoplankton Growth: Evidence from Biogeochemical Experiments and Satellite Data},'' \emph{Geophys. Res. Lett.}, vol.~34, no.~1, 2007.

\bibitem{chen2022federated}
X.~Chen, G.~Zhu, Y.~Deng, and Y.~Fang, ``{Federated Learning over Multihop Wireless Networks with In-network Aggregation},'' \emph{{IEEE} Trans. Wireless Commun.}, vol.~21, no.~6, pp. 4622--4634, 2022.

\bibitem{xie2019asynchronous}
C.~Xie, S.~Koyejo, and I.~Gupta, ``{Asynchronous Federated Optimization},'' \emph{arXiv preprint arXiv:1903.03934}, 2019.

\bibitem{zagoruyko2016wide}
S.~Zagoruyko and N.~Komodakis, ``Wide residual networks,'' \emph{arXiv preprint arXiv:1605.07146}, 2016.

\bibitem{tan2019efficientnet}
M.~Tan and Q.~Le, ``Efficientnet: Rethinking model scaling for convolutional neural networks,'' in \emph{Proc. ICML}, 2019, pp. 6105--6114.

\bibitem{lin2023split}
Z.~Lin, G.~Qu, X.~Chen, and K.~Huang, ``{Split Learning in 6G Edge Networks},'' \emph{IEEE Wirel. Commun.}, 2024.

\bibitem{lin2023pushing}
Z.~Lin, G.~Qu, Q.~Chen, X.~Chen, Z.~Chen, and K.~Huang, ``{Pushing Large Language Models to the 6G Edge: Vision, Challenges, and Opportunities},'' \emph{arXiv preprint arXiv:2309.16739}, 2023.

\bibitem{diaoheterofl}
E.~Diao, J.~Ding, and V.~Tarokh, ``{HeteroFL: Computation and Communication Efficient Federated Learning for Heterogeneous Clients},'' in \emph{Proc. ICLR}, 2021.

\bibitem{krizhevsky2012imagenet}
A.~Krizhevsky, I.~Sutskever, and G.~E. Hinton, ``{Imagenet Classification with Deep Convolutional Neural Networks},'' \emph{Proc. NIPS}, 2012.

\bibitem{helber2018introducing}
P.~Helber, B.~Bischke, A.~Dengel, and D.~Borth, ``{Introducing Eurosat: A Novel Dataset and Deep Learning Benchmark for Land Use and Land Cover Classification},'' in \emph{Proc. IGARSS}, 2018.

\bibitem{elmahallawy2022fedhap}
M.~Elmahallawy and T.~Luo, ``{FedHAP: Fast Federated Learning for LEO Constellations Using Collaborative HAPs},'' in \emph{Proc. WCSP}, 2022.

\end{thebibliography}

\end{document}